\documentclass[10pt,twocolumn,letterpaper]{article}
\usepackage{cvpr}
\usepackage{booktabs}  
\usepackage{multirow}  
\usepackage{graphicx}  
\usepackage[above]{placeins}  
\usepackage[T1]{fontenc}
\usepackage{amsmath,amssymb}
\usepackage{graphicx,booktabs,tabularx}
\usepackage{xcolor}
\usepackage{url}
\newcommand{\framework}{Real2Gym}

\newcommand{\blfootnote}[1]{%
  \begingroup
  \renewcommand{\thefootnote}{}%
  \footnotetext[0]{#1}%
  \endgroup
}

\usepackage{enumitem}

\usepackage{makecell,pdflscape,caption,listings,float,multirow}
\usepackage[normalem]{ulem}
\usepackage{xcolor}
\usepackage{listings}
\usepackage{tcolorbox}
\tcbuselibrary{skins,breakable,listings}

\definecolor{r2gPromptInk}{HTML}{285449}
\definecolor{r2gPromptBlue}{HTML}{326859}
\definecolor{r2gPromptViolet}{HTML}{806344}
\definecolor{r2gPromptPaper}{HTML}{FAFAF6}
\definecolor{r2gPromptTitle}{HTML}{E6EEE7}
\definecolor{r2gPromptBorder}{HTML}{D1DCD4}

\tcbset{r2gpromptpanel/.style={
  enhanced,breakable,lines before break=5,
  colback=r2gPromptPaper,colframe=r2gPromptBorder,
  colbacktitle=r2gPromptTitle,coltitle=r2gPromptInk,
  boxrule=0.35pt,arc=1pt,
  borderline west={1.5pt}{0pt}{r2gPromptInk},
  left=9pt,right=9pt,top=7pt,bottom=7pt,
  fonttitle=\normalfont\normalsize\bfseries,
  fontupper=\normalfont\normalsize\raggedright,
  before skip=9pt,after skip=11pt,
  before upper={\setlength{\parindent}{0pt}\setlength{\parskip}{2pt}}
}}
\newtcolorbox{rTwoGPrompt}[2]{
  r2gpromptpanel,title={#1},
  title after break={#1\enspace(continued)},
  before upper={\setlength{\parindent}{0pt}\setlength{\parskip}{2pt}%
    {\normalfont\normalsize\bfseries\color{r2gPromptViolet}#2}\par\smallskip}
}
\newcommand{\rTwoGPart}[1]{%
  \par\addvspace{3pt}\noindent
  {\color{r2gPromptBlue}\bfseries #1}\par\nobreak}

\lstdefinestyle{r2gappendixcode}{
  language={},basicstyle=\ttfamily\normalsize,
  columns=fullflexible,keepspaces=true,
  breaklines=true,breakatwhitespace=false,
  breakautoindent=false,breakindent=0pt,
  showstringspaces=false,showspaces=false,showtabs=false,
  numbers=none,frame=none,tabsize=2,
  aboveskip=0pt,belowskip=0pt,
  mathescape=false,texcl=false
}
\newtcblisting{rTwoGCode}[1]{
  r2gpromptpanel,title={#1},
  title after break={#1\enspace(continued)},
  listing only,listing options={style=r2gappendixcode,gobble=4}
}

\ifdefined\XeTeXversion
  \usepackage{newunicodechar}
  \newunicodechar{–}{--}
  \newunicodechar{—}{---}
\fi

\definecolor{cvprblue}{rgb}{0.21,0.49,0.74}
\usepackage[pagebackref,breaklinks,colorlinks,allcolors=cvprblue]{hyperref}
\def\confName{CVPR}
\def\confYear{2027}
\title{Real2Gym: Building Gyms from Videos, Bringing Skills to Robots}
\author{%
    Kerui Ren$^{1,2*}$\quad
    Yingxiang Xu$^{1,3*}$\quad
    Kaiwen Song$^{1,4}$\quad
    Linning Xu$^{5}$\\
    Mu Huang $^{6}$\quad
    Bo Dai$^{7}$\quad
    Mulin Yu$^{1\dagger}$\quad
    Tao Lu$^{1\dagger}$\\
  {\small
    $^{1}$Shanghai Artificial Intelligence Laboratory\quad
    $^{2}$Shanghai Jiao Tong University\quad
    $^{3}$Zhejiang University}\\
  {\small
    $^{4}$University of Science and Technology of China\quad
    $^{5}$The Chinese University of Hong Kong}\\
{\small
$^{6}$Fudan University\quad
    $^{7}$The University of Hong Kong}
}
\begin{document}
\makeatletter
\g@addto@macro\@maketitle{%
\begin{minipage}{\textwidth}
  \centering
  \includegraphics[width=\linewidth]{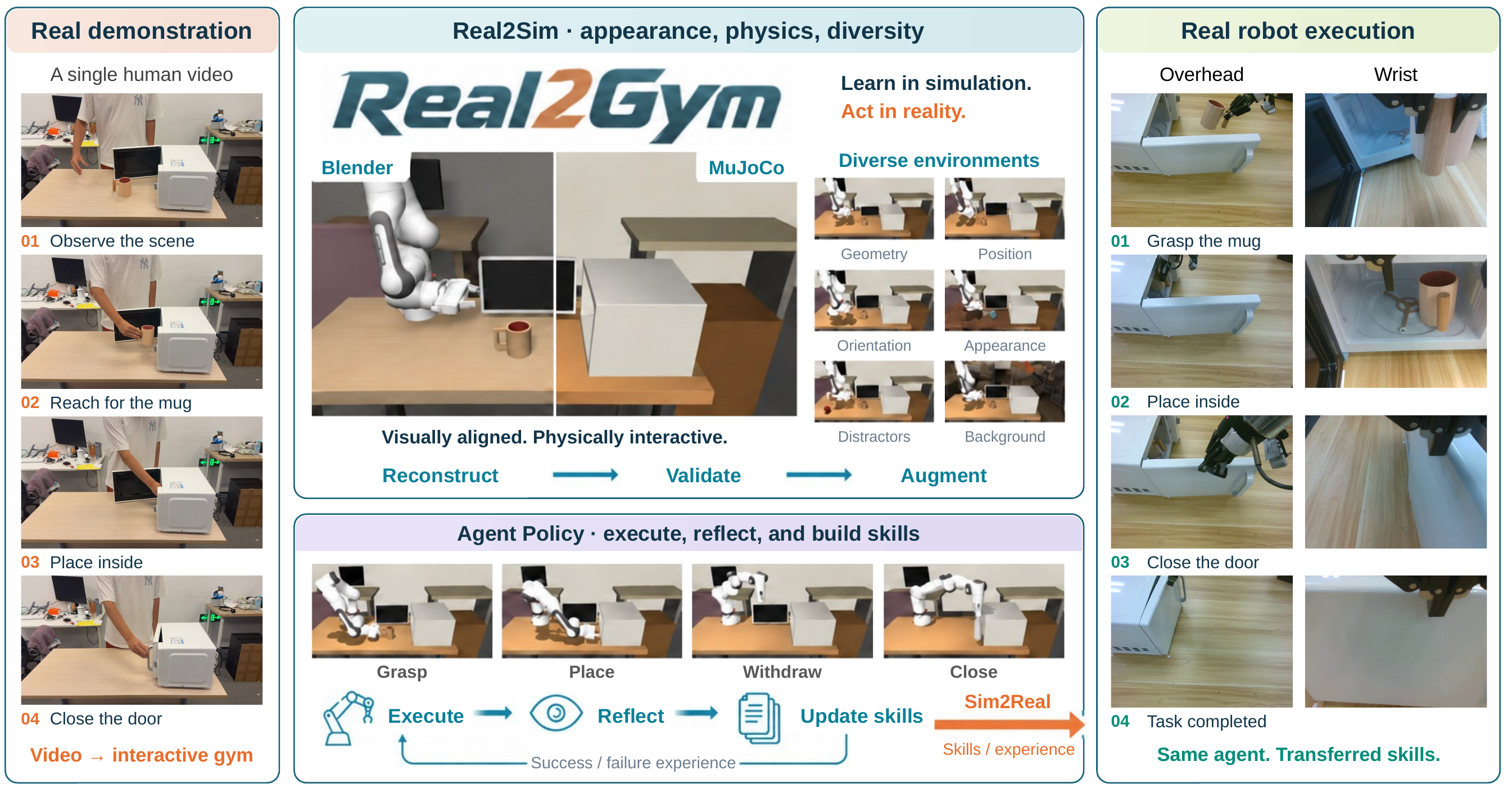}

  \captionof{figure}{
  Real2Gym converts human or robot videos into visually aligned and physically executable Blender and MuJoCo environments, where reconstructed interactions are validated under native physics and expanded into feasible task variations.
  Rather than treating simulation as the endpoint, the agent uses these gyms to execute, fail, reflect, and accumulate reusable skills. The resulting skills are re-grounded from human demonstrations and transferred to real robots through a shared control interface, enabling Real2Sim2Real self-improvement. Project page: 
\href{https://real2gym.github.io/}{\textcolor{magenta}{\textbf{https://real2gym.github.io/}}}.
  }
  \label{fig:teaser}
\end{minipage}\par\vspace{12pt}
}
\makeatother
\maketitle
\blfootnote{$^*$Equal contribution. $^\dagger$Corresponding author.}
\begin{abstract}
Real-world videos provide rich demonstrations of manipulation~\citep{droid,egodex}, but turning them into reusable robot skills requires visually aligned environments, executable physical interactions, and mechanisms for learning from experience~\citep{shinn2023reflexion,wang2023voyager}. We introduce \framework{}, an agentic Real2Sim2Real framework that turns human and robot demonstrations into interactive simulation gyms and brings skills acquired in simulation to physical robots. The Real2Sim module reconstructs editable scenes, aligns objects and cameras with the input, validates demonstrated or retargeted actions through native physics execution, and generates task-conditioned variations with action-feasibility checks. Within these environments, the agent generates executable code for manipulation stages, observes their outcomes, and distills successful attempts and failures into reusable task procedures, object-relative motions, and recovery strategies. Through a shared perception-and-control interface, these skills guide subsequent execution in simulation and on real robots, with motions adapted to current observations and no updates to the underlying model weights. Extensive evaluations demonstrate that \framework{} reconstructs high-fidelity simulation environments, where its policy with skills outperforms Agent as Policy by 33.3\% in success rate with approximately 44.2\% fewer policy-execution tokens. Through Real2Sim2Real experience, it improves real-world success on the same tasks by 55.0\% over execution without skills and generalizes to new scenes.
\end{abstract}

\section{Introduction}
\label{sec:introduction}

Robot self-improvement offers a promising pathway toward embodied intelligence: through iterative interaction, robots can diagnose execution failures, test policy adjustments, and accumulate experience to refine subsequent behavior~\citep{r2g_kober2013robotics}. Language agents likewise interleave reasoning with actions~\citep{yao2022react}, refine outputs using feedback~\citep{shinn2023reflexion,madaan2023self}, and accumulate executable skills for later reuse~\citep{wang2023voyager}. Recent robotic agents highlight how execution feedback and reusable skills drive this process~\citep{aspire2026,jia2026agent}. Despite this progress, deploying this trial-and-error paradigm directly on physical hardware remains prohibitively costly. Specifically, physical motion and scene resets are time-consuming, failed attempts risk damaging the manipulator or surrounding environment, and constant trials demand continuous supervision while accelerating hardware wear~\citep{r2g_dulac2019challenges}. Furthermore, limited physical setups restrict parallel exploration, and reliably restoring initial physical states after a failure is often difficult~\citep{r2g_eysenbach2017reset}. Consequently, scaling self-improvement entirely in the real world is bottlenecked by high operational costs, safety risks, and low sample efficiency.

To overcome these real-world bottlenecks, simulation offers a scalable setting for iterative trial-and-error before physical deployment~\citep{r2g_openai2018dexterity,r2g_tao2024maniskill3}. Modular simulators~\citep{zhu2020robosuite}, diverse household environments~\citep{nasiriany2024robocasa}, and lifelong-learning benchmarks~\citep{liu2023libero} support repeatable robot learning, while MimicGen scales demonstration data through object-relative trajectory adaptation~\citep{mandlekar2023mimicgen}. However, successful Real2Sim2Real transfer hinges on faithfully reproducing the target scene's task-relevant geometry, object articulations, and physical interactions~\citep{r2g_zhao2020sim2real}. Prior work addresses this transfer through visual randomization~\citep{r2g_tobin2017randomization} and physics calibration~\citep{r2g_tan2018sim2real}. Existing pipelines, such as RialTo~\citep{r2g_torne2024rialto}, construct these environments through a cumbersome workflow spanning scene scanning, mesh repair, articulation modeling, and physics parameterization. Heavily constrained by manual intervention and disjointed tools, building interactive simulation environments for new tasks remains prohibitively labor-intensive.

Addressing this heavy manual overhead, recent progress in visual geometry estimation~\citep{r2g_wang2023dust3r,r2g_wang2025vggt} and generative simulation~\citep{r2g_wang2023robogen} has accelerated automated scene construction. For instance, Agentic Real2Sim converts recordings of robot–object interactions into executable episodic twins~\citep{chen2026agenticreal2sim}, while state-of-the-art multimodal models like GPT-6 Astra can synthesize detailed Blender scenes from visual inputs. However, our comparisons show that GPT-6 Astra reconstructions can still contain inaccurate object dimensions, relative spatial layouts, and camera extrinsics (Fig.~\ref{fig:real2sim}). Errors in scale and placement distort grasp clearances and contact geometry, while camera misalignment obscures correspondence with the demonstrated motion. During fine-grained manipulation, these spatial inaccuracies cause interpenetration, missed contacts, and failed grasps, preventing the reconstructed scene from executing reliably under native physics. Consequently, such physical invalidity compromises the fidelity required for downstream agent exploration and skill accumulation.

We resolve these physical and geometric fidelity gaps with \framework{},  a unified Real2Sim2Real framework that couples physics-verified environment synthesis with agentic skill accumulation. Given a human or robot demonstration video, our Real2Sim pipeline constructs visually aligned Blender and MuJoCo environments, rigorously verifying physical interaction feasibility under native physics. Validated scenes are then augmented with action-adapted procedural variations to build diverse, execution-ready gyms for agent exploration. Within these environments, the agent operates across high-level manipulation stages, distilling execution feedback into reusable, object-relative skills that adapt to current observations without model weight updates. Across benchmark scenes reconstructed from public datasets, \framework{} markedly improves reconstruction quality. Its skill-guided agent achieves an $87.5\%$ success rate compared to $70.8\%$ for GPT-6 Astra Direct Mode, while consuming approximately $74.9\%$ fewer policy-execution tokens (Tab.~\ref{tab:real2sim} and Tab.~\ref{tab:agent_policy_average}). Finally, real-robot deployments confirm that simulation-refined skills enable successful physical execution on tasks that initially failed, validating our end-to-end pipeline.

Our main contributions are summarized as follows:
\begin{itemize}[leftmargin=1.25em]
    \item We introduce a Real2Sim pipeline that constructs \emph{executable} digital twins from human and robot demonstrations through event-centered correction, native-physics validation, and task-conditioned augmentation with action-feasibility checks.
    \item
    We introduce an experience-driven manipulation agent
    that integrates perception tools, executable operation stages, and feedback-driven skill extraction to support efficient interaction and skill reuse in simulation and on physical robots.
    \item 
    Across 24 reconstructed environments, \framework{} delivers consistent gains in reconstruction fidelity, task success, and token efficiency across both DROID and EgoDex, while real-robot experiments validate closed-loop Real2Sim2Real self-improvement.
    
\end{itemize}

\section{Related Work}
\label{sec:related}

\begin{figure*}[t]
\centering
\includegraphics[width=\linewidth]{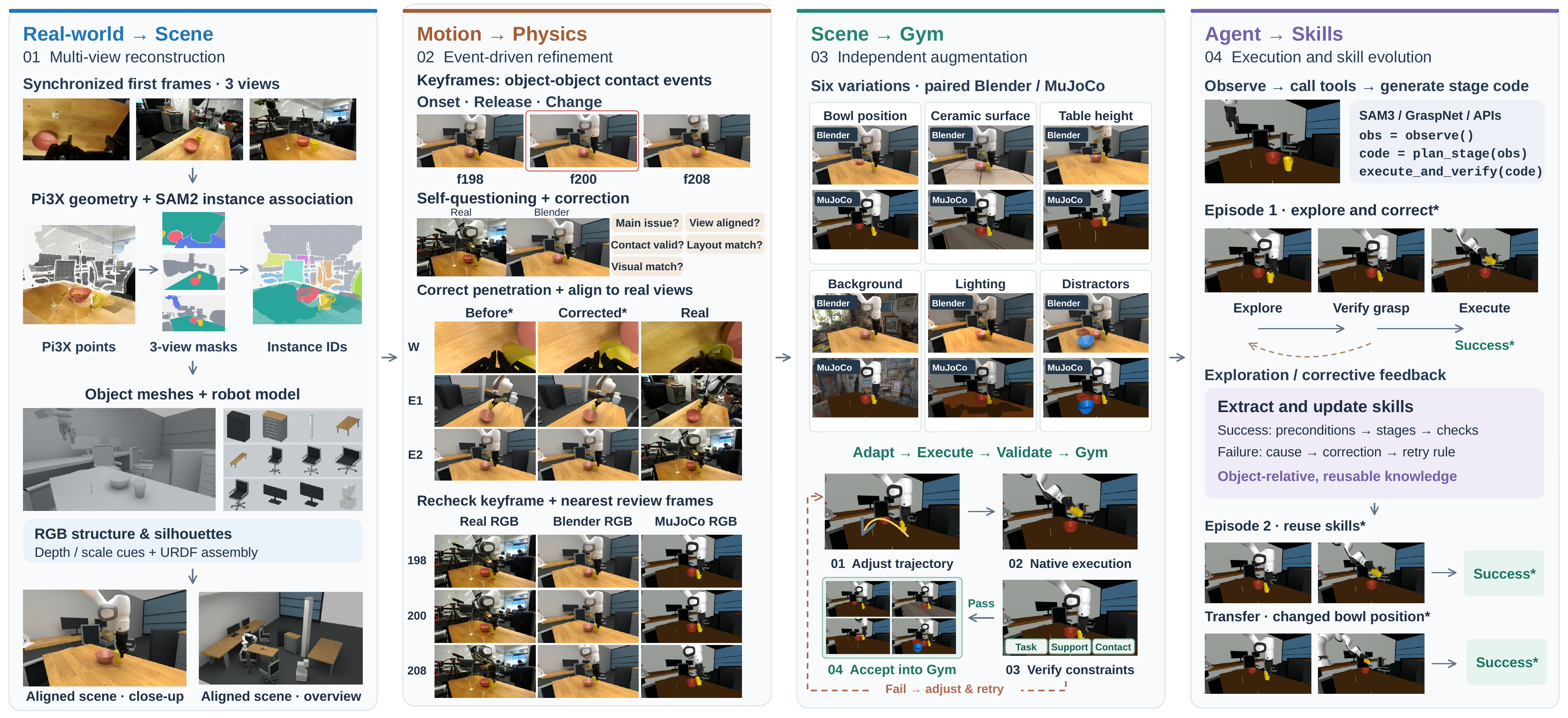}
\caption{\textbf{Overview of Real2Gym.} The reconstruction agent builds aligned Blender and MuJoCo scenes from human or robot demonstrations, refines them through event-driven correction and physics validation, and augments them into interactive gyms. The policy agent executes subtask-level code and distills feedback into reusable skills for simulation and real-robot deployment.}
\label{fig:overview}
\end{figure*}

\subsection{Real-to-Simulation Reconstruction}
Real-to-simulation reconstruction builds interactive digital twins from real observations and connects them to physics engines for robot policy training, data generation, and evaluation. Early frameworks such as RialTo~\citep{r2g_torne2024rialto} construct digital twins of real environments to refine manipulation policies via reinforcement learning before transferring them back to physical robots. Building on 3D Gaussian Splatting~\citep{r2g_kerbl2023gaussians}, subsequent methods integrate photorealistic observations with simulated interactions. Specifically, SplatSim~\citep{r2g_qureshi2024splatsim} leverages Gaussian rendering to generate RGB training data for zero-shot deployment, RoboGSim~\citep{r2g_li2024robogsim} couples Gaussian reconstruction with a physics engine for demonstration synthesis and closed-loop evaluation, and Splatting Physical Scenes~\citep{moran2025splatting} combines Gaussian appearances with explicit meshes to jointly refine geometry, robot poses, and physical parameters through differentiable rendering and MuJoCo simulation. More recently, agentic approaches automate the coordination of perception, modeling, and simulation tools, exemplified by Agentic Real2Sim~\citep{chen2026agenticreal2sim}, which uses vision-language agents to convert robot--object interaction recordings into executable episodic twins refined via simulator feedback.

\subsection{Agents for Robot Control}
Language-model-based robot control translates task instructions into executable behaviors through perception and control interfaces. Pioneering works ground high-level planning in skill affordances, as in SayCan~\citep{r2g_ahn2022saycan}, or synthesize programs for spatial reasoning and action composition, as in Code as Policies (CaP)~\citep{liang2023code}, while complementary methods close the perception-action loop via composable 3D value maps in VoxPoser~\citep{r2g_huang2023voxposer} and execution-driven replanning in Inner Monologue~\citep{r2g_huang2022inner}. Building on this foundation, recent coding-agent frameworks emphasize iterative self-correction and experience reuse: CaP-X~\citep{r2g_fu2026capx} benchmarks how multi-turn interaction and structured execution feedback enhance program refinement, whereas ASPIRE~\citep{aspire2026} combines evolutionary search with a persistent skill library distilled from validated code repairs. Extending this paradigm to general-purpose execution, Agent as Policy~\citep{jia2026agent} unifies visual interpretation, program generation, and physical adaptation without task-specific training, substantially accelerating repeated real-robot trials by reusing saved procedures.

\section{Method}
\label{sec:method}

Fig.~\ref{fig:overview} outlines \framework{}, which converts demonstrations into interactive environments and reusable manipulation skills through two distinct agents. Given a human or robot video $\mathcal{I}=\{\mathbf{I}_t^v\}_{t,v}$, indexed by time $t$ and view $v$, and a robot URDF $\mathcal{U}$, the reconstruction agent builds an environment pool $\mathcal{G}$ through scene construction, action reconstruction, and augmentation (Sec.~\ref{sec:real2sim}). Each environment pairs a visually aligned Blender scene with an interactive MuJoCo model. Within these environments, the policy agent $\pi_\theta$ executes tasks and accumulates a reusable skill library $\mathcal{K}$ from experience (Sec.~\ref{sec:agent}).

\subsection{Interactive Gym Construction}
\label{sec:real2sim}

\paragraph{Scene Reconstruction and Alignment.}
The reconstruction agent builds an editable scene that preserves the demonstrated objects, layout, and viewpoints. It initializes geometry from the first frame using MoGe-3~\citep{kong2026moge} for a single view or Pi3X~\citep{wang2026pi} across views, with calibrated cameras and metric cues anchoring scale and a shared coordinate frame. Semantic reasoning and SAM2~\citep{ravi2024sam2} segmentation identify manipulated objects, supports, and salient background entities with consistent identities across views. Their instance masks isolate point clouds that initialize object scale, position, and orientation. Guided by these estimates, RGB silhouettes, and visible structures, the agent selects geometric representations and generates editable modeling code. It completes unobserved regions using structural assumptions about thickness, closed back surfaces, and container walls, constrained by up to ten sparse subsequent RGB frames.

To check these meshes against the observations, the agent renders them from the input viewpoints and refines shape and pose based on silhouettes, occlusion ordering, and support relations. It then imports the target robot through its URDF or MJCF description and aligns its base pose, initial joints, and camera mount. Scene-level reprojection checks jointly refine object and robot alignment and camera parameters under support and kinematic constraints. This shared geometry provides the basis for recovering and validating the demonstrated interactions.

\paragraph{Action Reconstruction and Physical Validation.}
Building on the aligned scene, the agent uniformly samples candidate video frames and compares interaction relations between adjacent candidates. Changes in contact, grasp, support, or containment involving the hand or gripper, objects, and surrounding surfaces determine the representative event keyframes. For robot demonstrations, recorded joint and gripper states provide motion references when available. Human demonstrations provide only RGB video, without parameters of the MANO hand model~\citep{r2g_romero2017mano}, 3D hand poses, or hand trajectories. The agent infers hand interactions and object motion from this video and maps them to target-robot motion references.

At each event keyframe and available view, the agent checks five aspects: the primary discrepancy, camera alignment, relative object placement, contact or penetration, and appearance. Detected anomalies trigger neighboring-frame inspection, localized correction, and re-verification. Once visually aligned, the scene is instantiated in MuJoCo~\citep{r2g_todorov2012mujoco} with articulated robots, collision geometries, container cavities, physical materials, and actuators. The agent first calibrates approach trajectories and contact orientations, then checks gripper closure, grasp retention, support stability, and release. An end-to-end execution from the initial state validates physical consistency and task completion. The agent then reimports the native trajectory into Blender for frame-matched comparisons of real, Blender, and MuJoCo RGB images.

\paragraph{Task-Conditioned Scene Augmentation.}
Starting from a validated seed, the agent varies object geometry, pose, support height, materials, distractors, background, and lighting. Each factor is initially changed in isolation to assess task feasibility. Geometry changes propagate consistently to visual models, collision meshes, supports, and robot mounting constraints; manipulation stages are then adapted to the revised grasp regions, target poses, and clearances.

Each candidate undergoes native physics execution to check task completion, contact dynamics, support stability, and rendering consistency between Blender and MuJoCo. Failed candidates return for scene or action refinement, while validated candidates enter $\mathcal{G}$. This validation loop expands the environment pool while retaining only candidates that pass the physical checks.

\subsection{Agent Execution and Skill Accumulation}
\label{sec:agent}

\paragraph{Subtask-Level Closed-Loop Execution.}
Within the constructed environments, the policy agent alternates observation, code generation, execution, and feedback at the subtask level. At decision step $k$, it generates a Python program from task instruction $g$, current observations $o_k$, episode history $\mathcal H_k$, and explicitly selected skills $\mathcal K_e$:
\begin{equation}
    c_k = \pi_{\theta}(g, o_k, \mathcal H_k, \mathcal K_e).
\end{equation}
Observations include available camera views, robot proprioception, and permitted execution feedback. Each program $c_k$ defines entry conditions, intermediate checks, and an observable completion condition. It combines SAM3~\citep{carion2025sam3} segmentation and Contact-GraspNet~\citep{r2g_sundermeyer2021contactgraspnet} grasp proposals with coordinate transformations, target-pose offsets, and control commands such as \emph{goto\_pose()}, \emph{open\_gripper()}, and \emph{close\_gripper()}.

Conditional checks on updated observations let one response coordinate related actions, such as approach, closure, and grasp-retention testing. The executor runs the code in bounded segments and returns observations and feedback for the agent's next decision. Decisions share a continuous context within an episode; each new episode starts with a fresh context and its selected skills. Task success is evaluated independently from recorded execution under predefined criteria, providing outcome evidence for subsequent skill extraction.

\begin{table*}[t]
\centering
\caption{\textbf{Quantitative comparison of Real2Sim reconstruction.}
Rubric scores are reported in the order of GPT-6 Astra / GPT-6.1 Sol / Claude Opus 5.5, while LPIPS-VGG and DINOv2 patch cosine similarity (DINOv2 Cos.) measure visual alignment between the initial simulation render and the corresponding real video frame. }
\label{tab:real2sim}

\small
\setlength{\tabcolsep}{3pt}
\renewcommand{\arraystretch}{1.1}

\resizebox{\textwidth}{!}{%
\begin{tabular}{clcccccc}
    \toprule
    & \textbf{Method}
    & \textbf{Content Alignment} $\uparrow$
    & \textbf{Viewpoint Alignment} $\uparrow$
    & \textbf{Action Fidelity} $\uparrow$
    & \textbf{Simulation Success (/80)} $\uparrow$
    & \textbf{LPIPS-VGG} $\downarrow$
    & \textbf{DINOv2 Cos.} $\uparrow$ \tabularnewline
    \midrule
    \multirow{4}{*}{\rotatebox[origin=c]{90}{\textbf{DROID}}}
    & GPT-5.6 Sol
    & 46.080 / 46.420 / 48.170
    & 30.330 / 33.250 / 27.500
    & 69.170 / 71.830 / 63.670
    & 59.170 / 62.000 / 57.560
    & 0.740 & 0.174 \tabularnewline
    & GPT-6 Astra
    & 47.330 / 60.250 / 59.330
    & 30.500 / 36.500 / 31.670
    & 74.500 / 80.500 / 65.580
    & 67.500 / 69.670 / 55.920
    & 0.740 & 0.205 \tabularnewline
    & Agentic Real2Sim
    & 40.000 / 41.200 / 40.800
    & 28.900 / 29.500 / 31.000
    & 36.100 / 40.700 / 40.700
    & 29.600 / 32.670 / 24.670
    & 0.742 & 0.156 \tabularnewline
    & \textbf{Ours}
    & \textbf{81.250 / 69.080 / 73.330}
    & \textbf{82.670 / 86.670 / 80.750}
    & \textbf{89.500 / 89.920 / 82.920}
    & \textbf{76.250 / 73.250 / 67.830}
    & \textbf{0.617} & \textbf{0.364} \tabularnewline
    \midrule
    \multirow{4}{*}{\rotatebox[origin=c]{90}{\textbf{EgoDex}}}
    & GPT-5.6 Sol
    & 61.670 / 62.750 / 60.420
    & 41.420 / 43.500 / 43.330
    & 71.080 / 72.250 / 65.220
    & 59.250 / 60.830 / 59.330
    & 0.697 & 0.160 \tabularnewline
    & GPT-6 Astra
    & 63.170 / 62.830 / 63.920
    & 36.420 / 40.250 / 45.250
    & 71.750 / 73.420 / 67.250
    & 61.000 / 60.670 / 56.580
    & 0.703 & 0.181 \tabularnewline
    & Agentic Real2Sim
    & N/A & N/A & N/A & N/A & N/A & N/A \tabularnewline
    & \textbf{Ours}
    & \textbf{82.420 / 79.920 / 79.420}
    & \textbf{73.580 / 84.750 / 75.920}
    & \textbf{91.580 / 90.920 / 85.420}
    & \textbf{78.000 / 77.330 / 70.080}
    & \textbf{0.580} & \textbf{0.224} \tabularnewline
    \bottomrule
\end{tabular}%
}
\end{table*}

\paragraph{Extraction and Reuse.}
After each episode, a separate extraction process analyzes observations, generated code, feedback, and the final outcome. It assigns each code round a positive, negative, or unknown local effect, distinguishing successful substeps from unsuccessful attempts. Where the record supports comparison, it relates failed attempts to subsequent corrections. The extracted skills specify applicability conditions, task procedures, effect checks, object-relative motion rules, and recovery guidance. Each motion rule identifies the acting entity, an object anchor, a relative position or orientation, and a stopping condition.

During execution, these rules are instantiated using object poses estimated from current observations, allowing skills to adapt to changed object placements. New lessons are merged into the library with supporting evidence and revisions to contradicted guidance, then tested through fresh execution. The same representation supports real-robot deployment through the corresponding perception-and-control interface, which grounds skills in live observations. Experience thus updates the skill library while keeping model parameters $\theta$ fixed.

\section{Experiments}

\subsection{Experimental Setup}

\paragraph{Datasets \& Metrics.}
We select 12 scenes each from DROID~\citep{droid} and EgoDex~\citep{egodex} to evaluate our Real2Sim pipeline. DROID provides real-world robot demonstrations with external and wrist-mounted RGB videos, whereas EgoDex provides egocentric human manipulation videos. Each dataset contributes four easy, four medium, and four hard scenes. The 24 MuJoCo environments reconstructed by our pipeline subsequently serve as the simulation testbed for robotic agents. Notably, our pipeline uses only raw RGB videos without object trajectories or hand annotations. 

For Real2Sim evaluation, we adapt the LLM-as-a-judge paradigm~\citep{r2g_zheng2023judge} to a blinded, single-candidate protocol with an anonymized, evidence-grounded rubric. GPT-6 Astra high, GPT-6.1 Sol high, and Claude Opus 5.5 medium independently score each candidate across four dimensions: \textit{Content Alignment}, \textit{Viewpoint Alignment}, \textit{Action Fidelity}, and \textit{Simulation Success}. Additionally, we measure initial-frame visual correspondence between real videos and simulated renderings via LPIPS-VGG~\citep{r2g_zhang2018lpips} ($\downarrow$) and DINOv2~\citep{r2g_oquab2024dinov2} patch cosine similarity ($\uparrow$). Together, these metrics capture both visual fidelity to the source demonstration and the physical plausibility of simulated interactions. For robotic agent policy evaluation, we measure task success rate, response count, token usage, and execution time, where success is verified via predefined completion criteria and human inspection, while the remaining metrics quantify interaction efficiency.

\begin{figure}[t]
\centering
\includegraphics[width=\columnwidth]{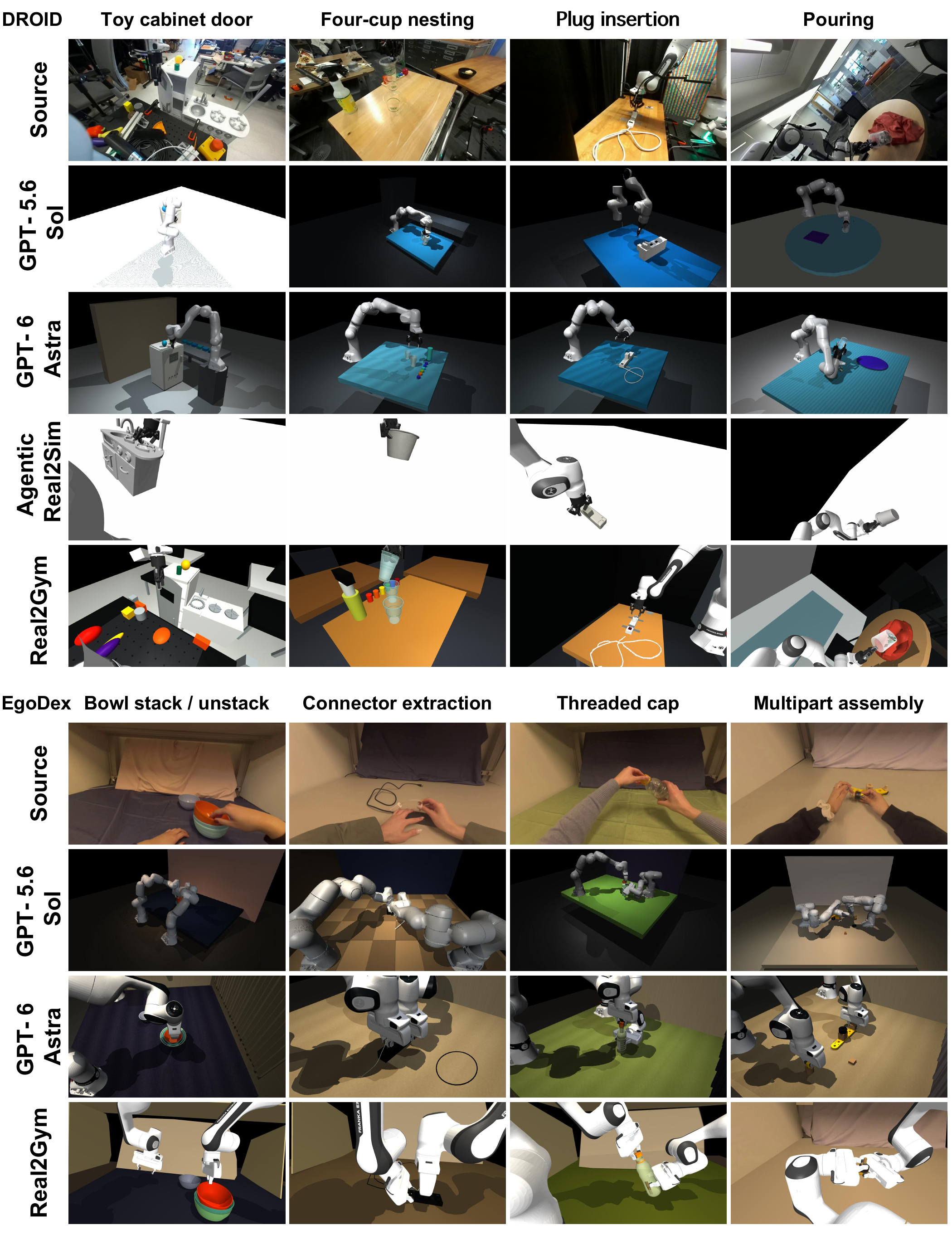}
\caption{\textbf{Qualitative reconstruction comparison on DROID (top) and EgoDex (bottom) tasks.} Our method achieves closer alignment with the reference inputs while maintaining the highest action consistency.}
\label{fig:real2sim}
\end{figure}

\paragraph{Real-World Hardware \& Task Suite.}
We evaluate our Real2Sim2Real pipeline on a 7-DoF Franka Emika Research 3 equipped with a Robotiq 2F-85 gripper and two Intel RealSense RGB-D cameras. One camera is mounted on the wrist, and the other provides a fixed external view. For each real-world scenario, we provide the system with a single human demonstration video, which serves as its only source of prior information regarding the scene and task. The system uses this demonstration to acquire and refine task skills in simulation before deploying them on the physical robot.

The task suite comprises four manipulation tasks:
(1) \textbf{Place Cup in Microwave and Close Door}, a sequential task involving the manipulation of articulated objects;
(2) \textbf{Toast Bread}, a sequential task requiring precise manipulation;
(3) \textbf{Hang the Mug}, a precision task in which the mug handle initially faces away from the camera, limiting its visibility; and
(4) \textbf{Place Plate on Shelf}, a challenging grasping and insertion task with tight spatial tolerances that requires delicately grasping the plate by its thin rim, aligning it with the target slot, and seating it securely in the narrow shelf.
We run 10 independent trials per method on each task.

\paragraph{Baselines.}
For Real2Sim evaluation, we compare our pipeline against Agentic Real2Sim~\citep{chen2026agenticreal2sim}, as well as GPT-6 Astra (medium) and GPT-5.6 Sol (xhigh), which reconstruct Blender and MuJoCo environments directly from demonstrations and robot models. For robotic policy evaluation, we compare against $\pi_{0.5}$-DROID~\citep{intelligence2025pi_}, Agent as Policy~\citep{jia2026agent}, and Direct Mode harnesses built on GPT-6 Astra (medium) and GPT-5.6 Sol (xhigh), which execute closed-loop actions from task instructions and live multi-view observations. Further implementation and configuration details are provided in the supplementary material.

\begin{table}[t]
\centering
\caption{\textbf{Quantitative comparison of agent policies.} SR is the success rate over 12 tasks per dataset. Resp., Tok., and Time (s) report mean model responses, tokens in millions, and execution time in seconds, including failures. \textbf{Bold} and \underline{underlined} denote best and second-best results.}
\label{tab:agent_policy_average}
\small
\setlength{\tabcolsep}{0.8pt}
\renewcommand{\arraystretch}{1.1}
\resizebox{\columnwidth}{!}{%
\begin{tabular}{lrrrr@{\hspace{5pt}}rrrr}
\toprule
& \multicolumn{4}{c}{\textbf{DROID}} & \multicolumn{4}{c}{\textbf{EgoDex}} \\
\cmidrule(lr){2-5} \cmidrule(lr){6-9}
\textbf{Method} & \textbf{SR}$\uparrow$ & \textbf{Resp.}$\downarrow$ & \textbf{Tok}$\downarrow$ & \textbf{Time (s)}$\downarrow$ & \textbf{SR}$\uparrow$ & \textbf{Resp.}$\downarrow$ & \textbf{Tok}$\downarrow$ & \textbf{Time (s)}$\downarrow$ \\
\midrule
$\pi_{0.5}$-DROID & 0.00 & N/A & N/A & N/A & 0.00 & N/A & N/A & N/A \\
GPT-5.6 Sol xhigh & 0.58 & 71.4 & 8.57 & 1543.2 & 0.42 & 105.5 & 13.63 & 1723.8 \\
GPT-6 Astra Medium & \underline{0.75} & 28.3 & 1.42 & \underline{468.6} & \underline{0.67} & 33.8 & 2.56 & 618.6 \\
Agent as Policy & \underline{0.67} & 15.2 & 0.84 & \underline{296.4} & \underline{0.50} & 20.9 & 0.79 & 256.8 \\
\textbf{Ours} & \underline{0.75} & \underline{14.6} & \underline{0.62} & 457.8 & \textbf{0.83} & \underline{13.4} & \underline{0.51} & \underline{459.6} \\
\textbf{Ours (w/ demos)} & \underline{0.75} & \underline{13.4} & \underline{0.71} & 499.2 & \textbf{0.75} & \underline{14.0} & \underline{0.83} & \underline{651.0} \\
\textbf{Ours (w/ skills)} & \textbf{0.92} & \textbf{12.8} & \textbf{0.46} & \textbf{367.8} & \textbf{0.92} & \textbf{14.3} & \textbf{0.45} & \textbf{373.8} \\
\bottomrule
\end{tabular}%
}
\end{table}

\paragraph{Implementation Details.}

Scene construction, policy execution, and skill extraction all use GPT-6 Astra with medium reasoning effort. Each new task starts in an independent session, while successive decisions within the task retain the conversation context. For all simulation experiments, both the agentic baselines and our framework are strictly restricted to the designated observations and are prohibited from accessing privileged simulator information, such as ground-truth object poses, trajectories, or other internal simulator states. Our standard policy budget is 50 model decisions per task and 600 control steps per code execution. A task is unsuccessful if its predefined completion criteria remain unmet when the budget is exhausted. We use Blender 4.5.3 LTS for scene construction and rendering, MuJoCo 3.3.7~\citep{r2g_todorov2012mujoco} for physics simulation, SAM3 0.1.0~\citep{carion2025sam3} for agent perception, and the PyTorch~\citep{r2g_paszke2019pytorch} implementation of Contact-GraspNet~\citep{r2g_sundermeyer2021contactgraspnet} for grasp proposals.

\begin{figure}[t]
    \centering
    \includegraphics[width=\linewidth]{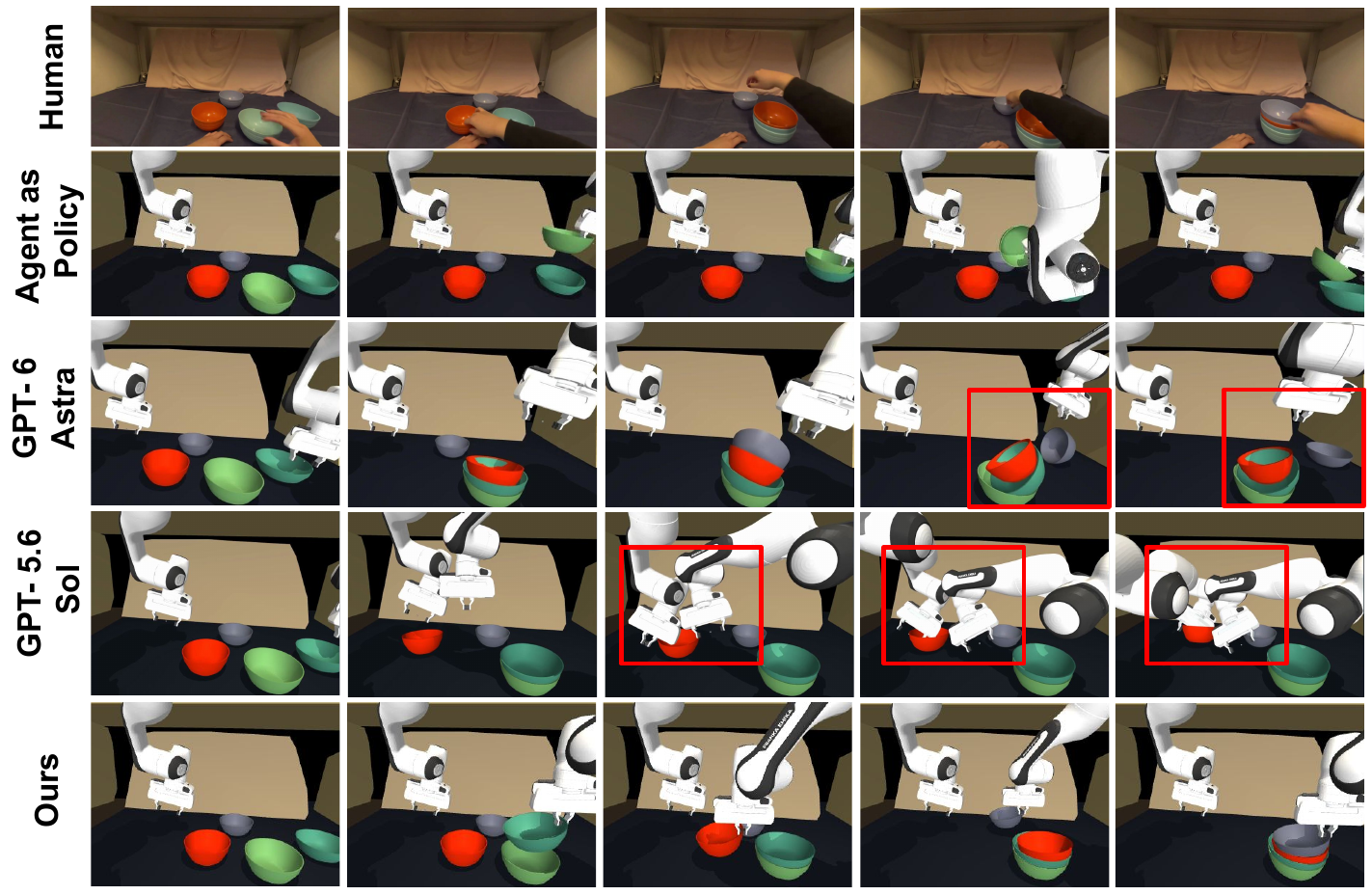}
    \caption{\textbf{Qualitative comparison of zero-shot agent execution.} Our agent successfully completes the stack, whereas both baselines fail, with red boxes highlighting key failure cases.}
    \label{fig:sim_robo_test}
\end{figure}

\begin{figure}[t]
    \centering
    \includegraphics[width=\linewidth]{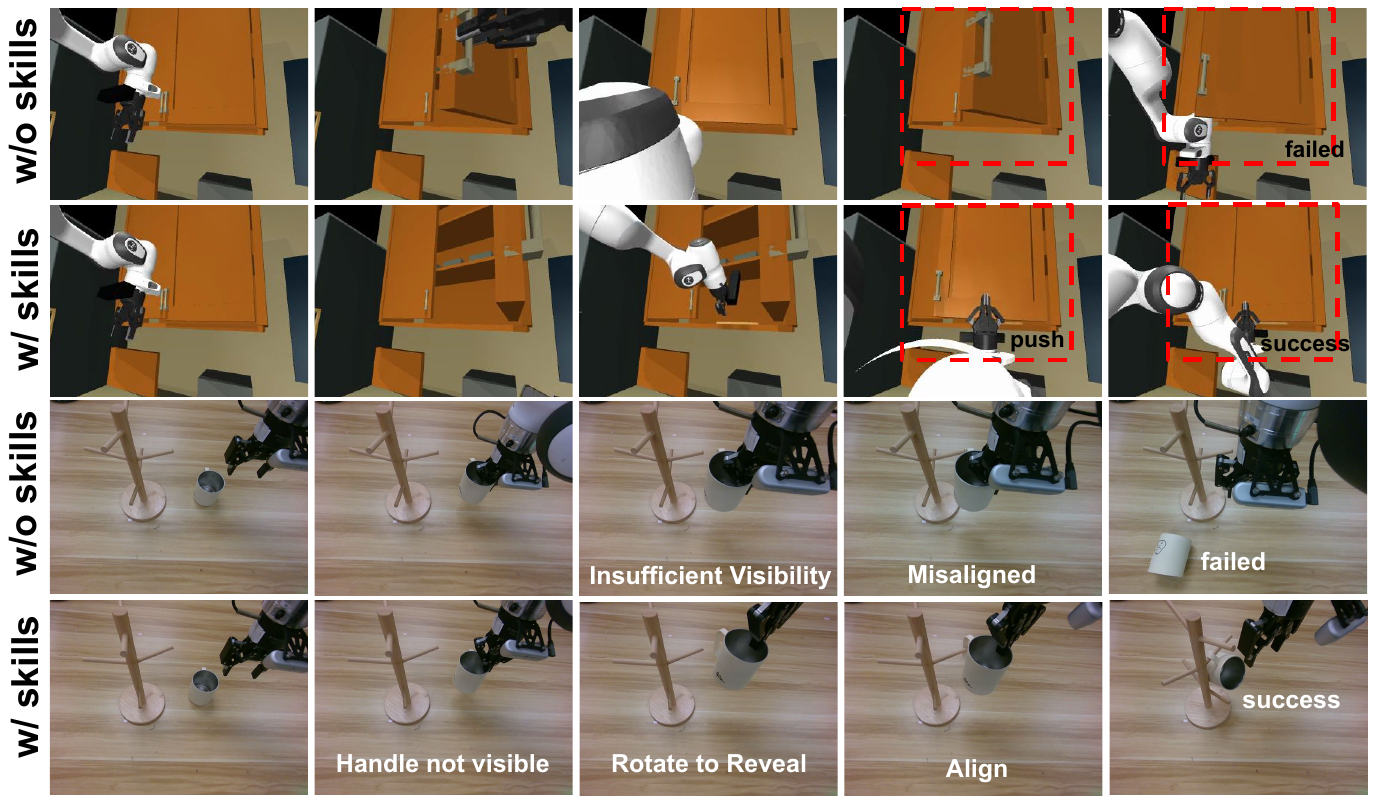}
    \caption{\textbf{Qualitative evaluation of skill reuse in simulation and real-world environments.} We compare robotic manipulation executions with and without skill reuse to demonstrate the effectiveness of reusable skills across simulated and real-world tasks.}
    \label{fig:skill_reuse}
\end{figure}

\subsection{Comparison}
\paragraph{Real2Sim Reconstruction.}
Tab.~\ref{tab:real2sim} shows that Real2Gym leads the reported baselines in both visual correspondence and interaction reconstruction, with consistent gains across the VLM judges. On DROID, Viewpoint Alignment exceeds GPT-6 Astra by 52.17 points under the GPT-6 Astra judge, while LPIPS-VGG is reduced by 16.6\% relative to the same baseline. Fig.~\ref{fig:real2sim} makes these gains concrete: Real2Gym better preserves the toy cabinet's structure, the cup arrangement, and the framing of the plug and cable, whereas baseline reconstructions often omit surrounding objects or shift the viewpoint. These improvements are consistent with instance-level geometric initialization followed by reprojection-based refinement, which jointly corrects object scale, placement, and camera alignment.

The gains extend to demonstrated interactions. On EgoDex, the Simulation Success score improves by 17.00 points over GPT-6 Astra under the same judge. In Fig.~\ref{fig:real2sim}, the human-demonstration reconstructions retain the nested-bowl arrangement and the relative placement of the bottle and cap while replacing human hands with robot grippers. Such task-relevant relations connect scene reconstruction to action recovery: event keyframes focus correction on changes in contact, support, and containment, while native physics checks test grasp retention, stable support, and release. The higher Action Fidelity and Simulation Success scores are therefore consistent with combining visual correction and physical validation, beyond the initial-frame appearance measured by LPIPS and DINOv2.

\paragraph{Agent Policy Performance in Simulation.}
Tab.~\ref{tab:agent_policy_average} shows that the policy agent improves task reliability with lower model interaction costs. Without accumulated skills, it matches GPT-6 Astra on DROID and raises EgoDex SR from 0.67 to 0.83, while using 71.6\% fewer tokens averaged equally across the datasets. This efficiency is consistent with subtask-level code that coordinates related actions and intermediate checks within one response. Fig.~\ref{fig:sim_robo_test} illustrates why these checks matter: GPT-6 Astra leaves the final bowl outside the stack, and GPT-5.6 Sol stalls during grasping, whereas our agent completes the stack. The sequence highlights the need to coordinate grasping, alignment, and release around the evolving object configuration, as implemented by the policy agent's loop of observation, execution, and feedback.

Skill reuse further raises mean SR from 0.79 to 0.92 across the datasets, whereas directly supplying demonstrations does not yield the same benefit. Fig.~\ref{fig:skill_reuse} provides a concrete example: the run without skills leaves the cabinet door ajar, while the skill-conditioned run applies a final push and completes closure. This correction illustrates the value of extracting object-relative motion rules and explicit effect checks from execution feedback, giving subsequent runs actionable guidance for unfinished subtasks. Agent as Policy remains faster, so the results support improved reliability and token economy without implying uniformly lower execution time.

\subsection{Real-World Deployment and Self-Evolution}

\begin{figure*}[t]
  \centering
  \includegraphics[width=\linewidth]{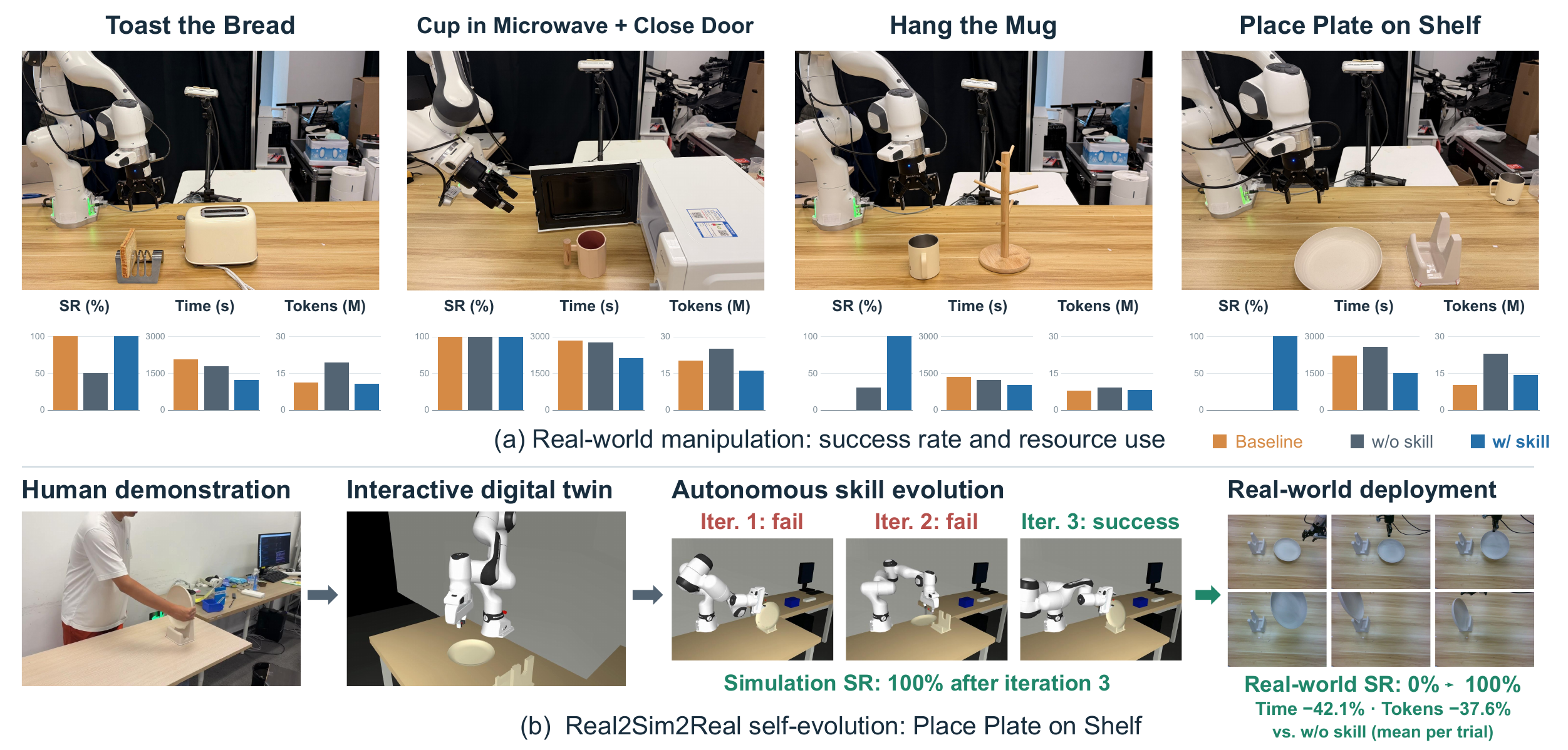}
  \caption{\textbf{Real-world deployment and Real2Sim2Real self-evolution.}
  (a) Our method outperforms Direct Mode in zero-shot success rates across four real-world manipulation tasks, while skill accumulation further contributes to task success.
  (b) Failure-driven self-evolution: reconstructing a digital twin from
  a human demonstration, acquiring skills through simulation, and
  deploying the evolved skills to improve real-world success.}
  \label{fig:real2gym_franka}
\end{figure*}

\begin{table}[t]
\centering
\caption{\textbf{Quantitative comparison on real-world robotic tasks.}
We compare the baseline, our method without skills (w/o skill), and our full
method with skill reuse (w/ skill) across four tasks.}
\label{tab:real_world_results}

\small
\setlength{\tabcolsep}{0.8pt}
\renewcommand{\arraystretch}{1.1}
\resizebox{\columnwidth}{!}{%
\begin{tabular}{lrrrr@{\hspace{5pt}}rrrr}
\toprule
& \multicolumn{4}{c}{\textbf{Toast Bread}} & \multicolumn{4}{c}{\textbf{Place Cup in Microwave}} \\
\cmidrule(lr){2-5} \cmidrule(lr){6-9}
\textbf{Method}
& \textbf{SR}$\uparrow$ & \textbf{Resp.}$\downarrow$ & \textbf{Tok.}$\downarrow$ & \textbf{Time (s)}$\downarrow$
& \textbf{SR}$\uparrow$ & \textbf{Resp.}$\downarrow$ & \textbf{Tok.}$\downarrow$ & \textbf{Time (s)}$\downarrow$ \\
\midrule
GPT-6 Astra Medium & \textbf{1.00} & 154.0 & 11.24 & 2064.0 & \textbf{1.00} & \textbf{201.0} & 20.28 & 2843.0 \\
Ours (w/o skill) & 0.50 & 196.0 & 19.30 & 1771.0 & \textbf{1.00} & 290.0 & 25.12 & 2763.0 \\
\textbf{Ours (w/ skill)} & \textbf{1.00} & \textbf{146.0} & \textbf{10.69} & \textbf{1223.0} & \textbf{1.00} & 271.0 & \textbf{16.00} & \textbf{2115.0} \\
\midrule
& \multicolumn{4}{c}{\textbf{Hang the Mug}} & \multicolumn{4}{c}{\textbf{Place Plate on Shelf}} \\
\cmidrule(lr){2-5} \cmidrule(lr){6-9}
\textbf{Method}
& \textbf{SR}$\uparrow$ & \textbf{Resp.}$\downarrow$ & \textbf{Tok.}$\downarrow$ & \textbf{Time (s)}$\downarrow$
& \textbf{SR}$\uparrow$ & \textbf{Resp.}$\downarrow$ & \textbf{Tok.}$\downarrow$ & \textbf{Time (s)}$\downarrow$ \\
\midrule
GPT-6 Astra Medium & 0.00 & 116.0 & \textbf{7.71} & 1337.0 & 0.00 & \textbf{148.0} & \textbf{10.16} & 2210.0 \\
Ours (w/o skill) & 0.30 & 127.0 & 9.11 & 1210.0 & 0.00 & 253.0 & 22.89 & 2586.0 \\
\textbf{Ours (w/ skill)} & \textbf{1.00} & \textbf{107.0} & 8.08 & \textbf{1005.0} & \textbf{1.00} & 176.0 & 14.28 & \textbf{1498.0} \\
\bottomrule
\end{tabular}%
}
\end{table}

Tab.~\ref{tab:real_world_results} shows that skill reuse improves both reliability and execution efficiency on physical hardware. Across the tasks, skill reuse raises mean SR from 0.45 to 1.00 compared with Ours (w/o skill). It also reduces mean execution time by 30.9\% compared with GPT-6 Astra. The task-level pattern in Fig.~\ref{fig:real2gym_franka}(a) distinguishes two benefits. For \textbf{Toast Bread} and \textbf{Place Cup in Microwave}, the baseline already completes the tasks, so skill reuse primarily improves efficiency relative to it. For \textbf{Hang the Mug} and \textbf{Place Plate on Shelf}, the main benefit is overcoming failures involving occluded geometry and constrained placement. The strongest reliability gains therefore occur where execution requires resolving task-specific geometric constraints.

The \textbf{Hang the Mug} example shows how an execution failure becomes actionable guidance. Skill reuse raises SR from 0.30 to 1.00 relative to the same policy without skills. In Fig.~\ref{fig:skill_reuse}, the unconditioned run approaches the rack with the handle insufficiently visible, misaligns the mug, and drops it. The skill-conditioned run first rotates the mug to reveal the handle, then aligns it with the rack and completes hanging. This sequence highlights the role of observation-dependent procedures: locating the handle is a prerequisite for selecting the approach, and successful placement must be checked before release. These requirements match the applicability conditions, object-relative motion rules, and effect checks extracted by the method. They explain why reusable guidance must specify what to observe and verify, alongside how to move.

\textbf{Place Plate on Shelf} further illustrates the feedback loop of Real2Sim2Real. Skill reuse raises SR from 0\% to 100\% over Ours (w/o skill) on \textbf{Place Plate on Shelf}. Fig.~\ref{fig:real2gym_franka}(b) traces the process from a human demonstration to a reconstructed environment, unsuccessful simulated attempts, a successful placement, and physical deployment. The thin-rim grasp and narrow shelf clearance couple grasp orientation with insertion alignment, making successful execution sensitive to intermediate states. The method retains corrections as object-relative rules and completion checks, then grounds them in live observations on the real robot. This representation provides a mechanism for transferring lessons from simulated failures to physical execution. GPT-6 Astra uses fewer tokens on these difficult tasks but does not complete them, illustrating why resource use must be interpreted together with task completion.





\section{Limitations}
\label{sec:limitations}
Although Real2Gym enables complex simulation-based exploration and successful Real2Sim2Real skill transfer, several limitations remain. First, regarding 3D reconstruction fidelity, multi-view misalignment between wrist-mounted and static cameras can yield ghosting or layered surfaces in noisy Pi3X point clouds, leaving residual geometric and textural discrepancies from the original source video. Enhancing cross-view registration and appearance refinement remains a key future priority. Second, regarding control granularity, while coarse stage-level code generation optimizes interaction efficiency, it lacks the local fine-grained reactivity required for highly constrained manipulation. We plan to explore adaptive switching between stage-level and fine action-level control to balance execution speed with local kinematic precision. Third, regarding embodiment coverage, our real-world experiments are currently confined to physical Franka arms with parallel-jaw grippers. Extending the framework to dexterous hands and diverse humanoid platforms will be critical to rigorously assessing its broader practical generality.

\section{Conclusion}
\label{sec:conclusion}

We introduce Real2Gym, a unified framework that converts human and robot demonstrations into interactive simulation environments and uses execution feedback to accumulate reusable robot skills. Our Real2Sim pipeline produces reconstructions with higher visual fidelity and more faithful physical interactions than the evaluated baselines. In simulation, our agent with accumulated skills achieves 91.7\% task success with approximately 77\% fewer policy-execution tokens than GPT-6 Astra. Experiments on a physical Franka robot also show higher overall success than direct GPT-6 Astra control, while skills refined in simulation enable a previously failed plate-placement task, completing the Real2Sim2Real loop. The framework supports building collections of simulation environments from human demonstrations, where agents make decisions, receive feedback, and update skills that subsequently guide physical robot execution. This provides a paradigm for robot self-improvement from real-world data, using simulation to turn human manipulation experience into reusable robot capabilities.

\subsection*{LLM Usage Statement}
\label{sec:statement}

In this work, we used generative AI tools to assist with generating
synthetic datasets, implementing methodologies and writing or modifying
research code, translating research-related text, improving the
readability and presentation of the manuscript, and searching for and
summarizing relevant literature. We did not use generative AI tools to
design the research methods or experiments, interpret experimental
results, propose or refine hypotheses, clean or reformat datasets,
support qualitative or thematic data analysis, formulate mathematical
claims, provide key elements for proving mathematical claims, or assist
in writing mathematical proofs; these tasks were either performed by the
authors or were not applicable to this work.

All AI-assisted work was carefully reviewed by the authors. Synthetic
datasets were inspected for relevance and correctness before use.
AI-assisted code was inspected, tested, and verified by the authors.
Translated and polished text was cross-checked sentence-by-sentence to
ensure that the original meaning was preserved. Literature identified or
summarized with the assistance of generative AI was independently checked
against the corresponding original sources. The authors take full
responsibility for the final content of this work, including all text,
methods, code, datasets, claims, experimental results, and other
artifacts produced with the assistance of generative AI.

{
    \small
    \bibliographystyle{ieeenat_fullname}
    \bibliography{main}
}

\clearpage
\appendix
\begingroup
The supplementary material is organized as follows. Sec.~\ref{app:reconstruction_ablation} presents a reconstruction ablation study, while Sec.~\ref{app:real2sim2real} details Real2Sim2Real validation and skill utilization. Sec.~\ref{app:model_effort_ablation} analyzes the impact of VLM model choices and reasoning effort. Additionally, Sec.~\ref{app:real2sim_results} and Sec.~\ref{app:simulation_results} report detailed per-task results for Real2Sim and simulation control, respectively. Finally, Sec.~\ref{app:evaluation_prompt} through Sec.~\ref{app:agent_policy_baseline} provide the Real2Sim evaluation criteria, baseline implementation details, and Agent Policy instructions.

\begin{figure*}[t]
  \centering
  \includegraphics[width=\textwidth]{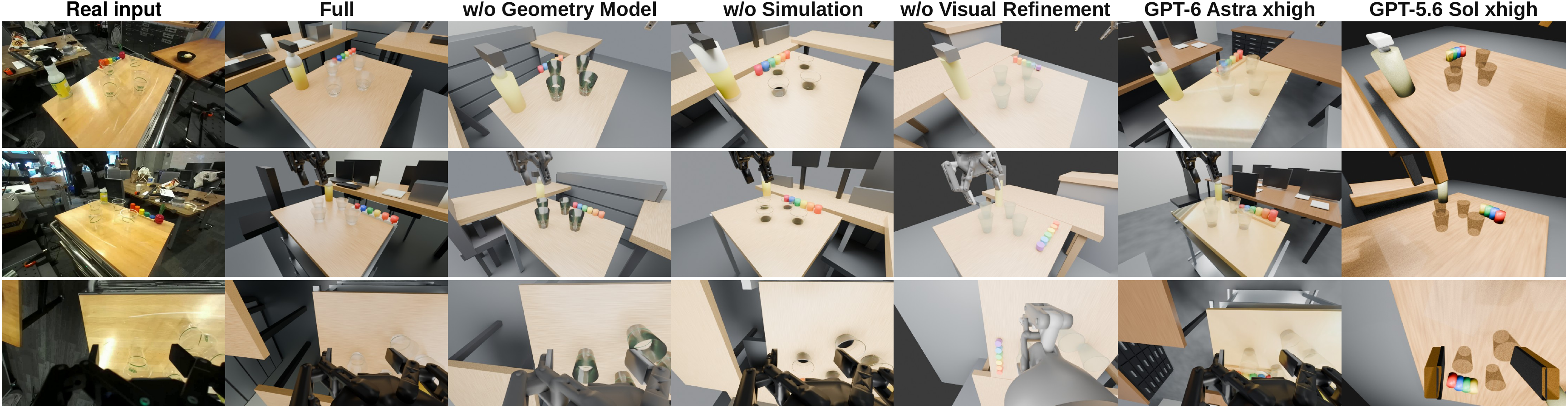}
  \caption{\textbf{First-frame reconstruction variants on D09.}
  Rows show three input camera viewpoints, and columns compare the real
  input with six Blender reconstructions.}
  \label{fig:d09_ablation}
\end{figure*}

\section{Ablation on Real2Sim Components}
\label{app:reconstruction_ablation}

\begin{table}[b]
\centering
\caption{\textbf{Comparison of reconstruction variants.}
All six conditions are evaluated on the same four-cup nesting scene.
LPIPS-VGG~\citep{r2g_zhang2018lpips} and DINOv2~\citep{r2g_oquab2024dinov2} patch cosine similarity compare the first frame
of each MuJoCo and Blender render with the corresponding frame from
the source video.}
\label{tab:d09_ablation}

\small
\setlength{\tabcolsep}{0.5pt}
\renewcommand{\arraystretch}{1.1}

\resizebox{\linewidth}{!}{%
\begin{tabular}{lcccc}
    \toprule
    & \multicolumn{2}{c}{\textbf{MuJoCo}}
    & \multicolumn{2}{c}{\textbf{Blender}} \tabularnewline
    \cmidrule(lr){2-3}\cmidrule(lr){4-5}
    \textbf{Condition}
    & \textbf{LPIPS-VGG} $\downarrow$
    & \textbf{DINOv2 Cos.} $\uparrow$
    & \textbf{LPIPS-VGG} $\downarrow$
    & \textbf{DINOv2 Cos.} $\uparrow$ \tabularnewline

    \midrule
    Full
    & 0.682 & 0.382 & 0.665 & 0.386 \tabularnewline
    \midrule
    w/o Visual Refinement
    & 0.730 & 0.304 & 0.748 & 0.299 \tabularnewline
    
    w/o Geometry Model
    & 0.657 & 0.356 & 0.706 & 0.347 \tabularnewline
    w/o Simulation Feedback
    & 0.701 & 0.349 & 0.707 & 0.361 \tabularnewline
        \midrule
    GPT-5.6 Sol (xhigh)
    & 0.775 & 0.225 & 0.730 & 0.234 \tabularnewline
    GPT-6 Astra (xhigh)
    & 0.627 & 0.426
    & 0.641 & 0.431 \tabularnewline
    \bottomrule
\end{tabular}%
}
\end{table}

Tab.~\ref{tab:d09_ablation} and Fig.~\ref{fig:d09_ablation} isolate three
Real2Sim components and reconstruction model choice on the D09 four-cup nesting
scene, using first-frame LPIPS-VGG and DINOv2 similarity in both MuJoCo and
Blender together with three-view qualitative comparisons. \textbf{(1) Visual
Refinement.} Removing it causes the largest component-level degradation:
Blender LPIPS-VGG increases from 0.665 to 0.748 and DINOv2 similarity decreases
from 0.386 to 0.299, while cup boundaries fade and the layout and camera
composition depart from the input views. This supports the render-and-correct
loop for refining shape, pose, and camera parameters. \textbf{(2) Geometry
Model.} Without explicit geometry modeling, DINOv2 similarity falls from 0.382
to 0.356 in MuJoCo and from 0.386 to 0.347 in Blender, and the cups acquire
dark, distorted profiles. Its slightly lower MuJoCo LPIPS-VGG, 0.657 versus
0.682, also shows that LPIPS-VGG alone can miss task-relevant structural
errors. \textbf{(3) Simulation Feedback.} Disabling physical feedback worsens
all four metrics; in Blender, LPIPS-VGG increases from 0.665 to 0.707 and
DINOv2 similarity decreases from 0.386 to 0.361, alongside larger deviations
in bottle pose and cup appearance. This is consistent with native physics
checks correcting contact- and support-inconsistent scenes. \textbf{(4) Model
Capacity.} Although GPT-6 Astra (xhigh) achieves the best scores across all four metrics and preserves the transparent cups, object layout, and surrounding furniture more faithfully than GPT-6 Astra (medium) and GPT-5.6 Sol (xhigh), this gain comes at the cost of a 183.6\% increase in cumulative token usage.

\section{Real2Sim2Real and Skill Generalization}
\label{app:real2sim2real}

This section provides additional qualitative evidence for the complete
Real2Sim2Real pipeline, from human demonstrations and simulation scene
construction to skill evolution, real-world execution, and generalization
to unseen configurations.

\subsection{Real2Sim Environment Reconstruction}
\label{app:real2sim_reconstruction}

\begin{figure*}[t!]
  \centering
  \includegraphics[width=0.98\textwidth,height=0.75\textheight,keepaspectratio]{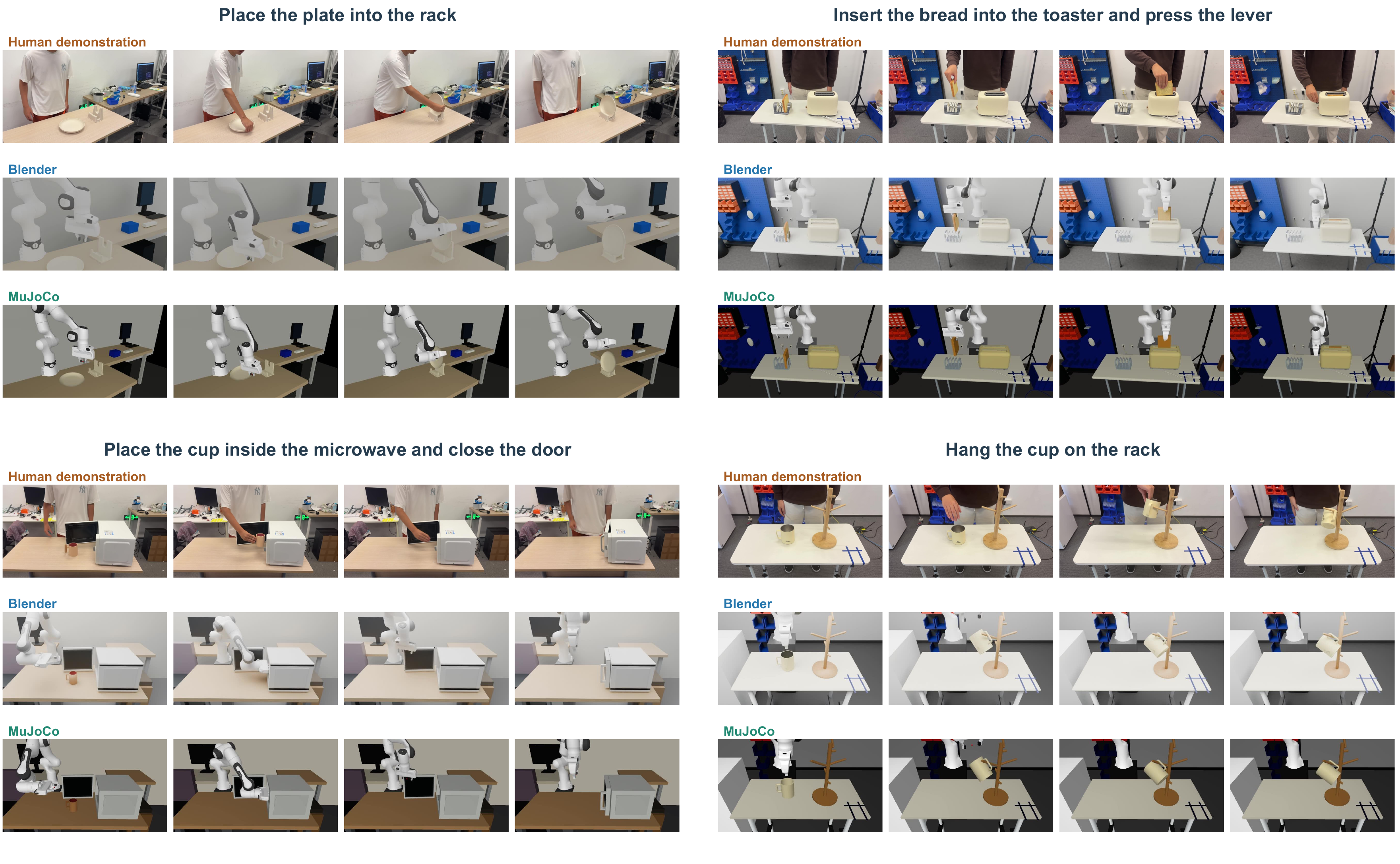}
  \caption{\textbf{Real2Sim environment reconstruction from human demonstrations.}
  Representative frames from four manipulation tasks (plate to rack, bread
  to toaster, cup to microwave, and mug to rack) are shown alongside their
  corresponding Blender reconstructions and MuJoCo environments.}
  \label{fig:supp_real2sim}
\end{figure*}

Fig.~\ref{fig:supp_real2sim} illustrates the transition from human
manipulation demonstrations to simulation environments across four tasks:
place a plate on a rack, toasting bread, place a cup in a microwave and close the door,
and hang a mug on a rack. Starting from task demonstrations, our
pipeline constructs the corresponding scenes in Blender and prepares
MuJoCo environments for physics-based interaction. The reconstructed
scenes preserve task-relevant object arrangements and spatial relations,
allowing the agent to explore manipulation strategies in simulation
before deploying them on the physical robot. This intermediate stage
supports the acquisition and refinement of task-relevant skills without
requiring each exploratory attempt to be performed on real hardware.

\subsection{Real-World Execution and Skill Transfer}
\label{app:real_world_sequences}

\begin{figure*}[t!]
  \centering
  \includegraphics[height=0.9\textheight,keepaspectratio]{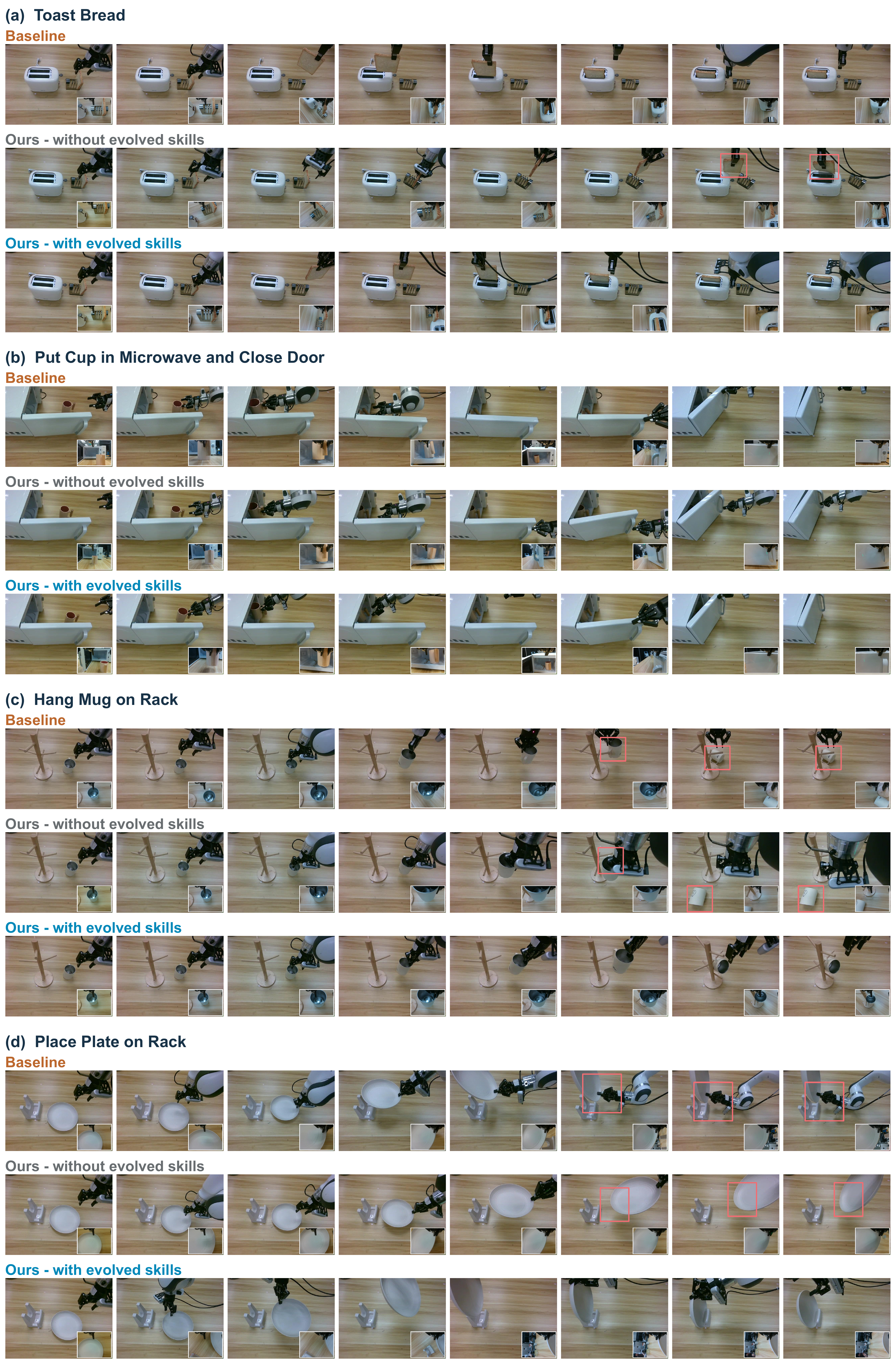}
  \caption{\textbf{Real-world manipulation trajectories.}
  Qualitative execution sequences for four tasks: (a) toast bread,
  (b) put a cup in a microwave and close the door, (c) hang a mug on a
  rack, and (d) place a plate on a rack. For each task, we compare the
  baseline, our agent without evolved skills, and our agent with evolved
  skills. Wrist-view insets highlight task-critical interactions; red
  boxes mark representative failure cases.}
  \label{fig:supplementary_franka}
\end{figure*}

Fig.~\ref{fig:supplementary_franka} supplements the quantitative
real-world evaluation with representative execution sequences of the
baseline, our agent without evolved skills, and our agent with evolved
skills. Across the four tasks, the skill-enhanced
agent exhibits more reliable execution, with especially pronounced
improvements on the two precision-sensitive tasks: hanging a mug on a
rack and placing a plate on a rack.

\subsection{Generalization of Evolved Skills}
\label{app:skill_generalization}

\begin{figure*}[t!]
  \centering
  \includegraphics[width=\textwidth]{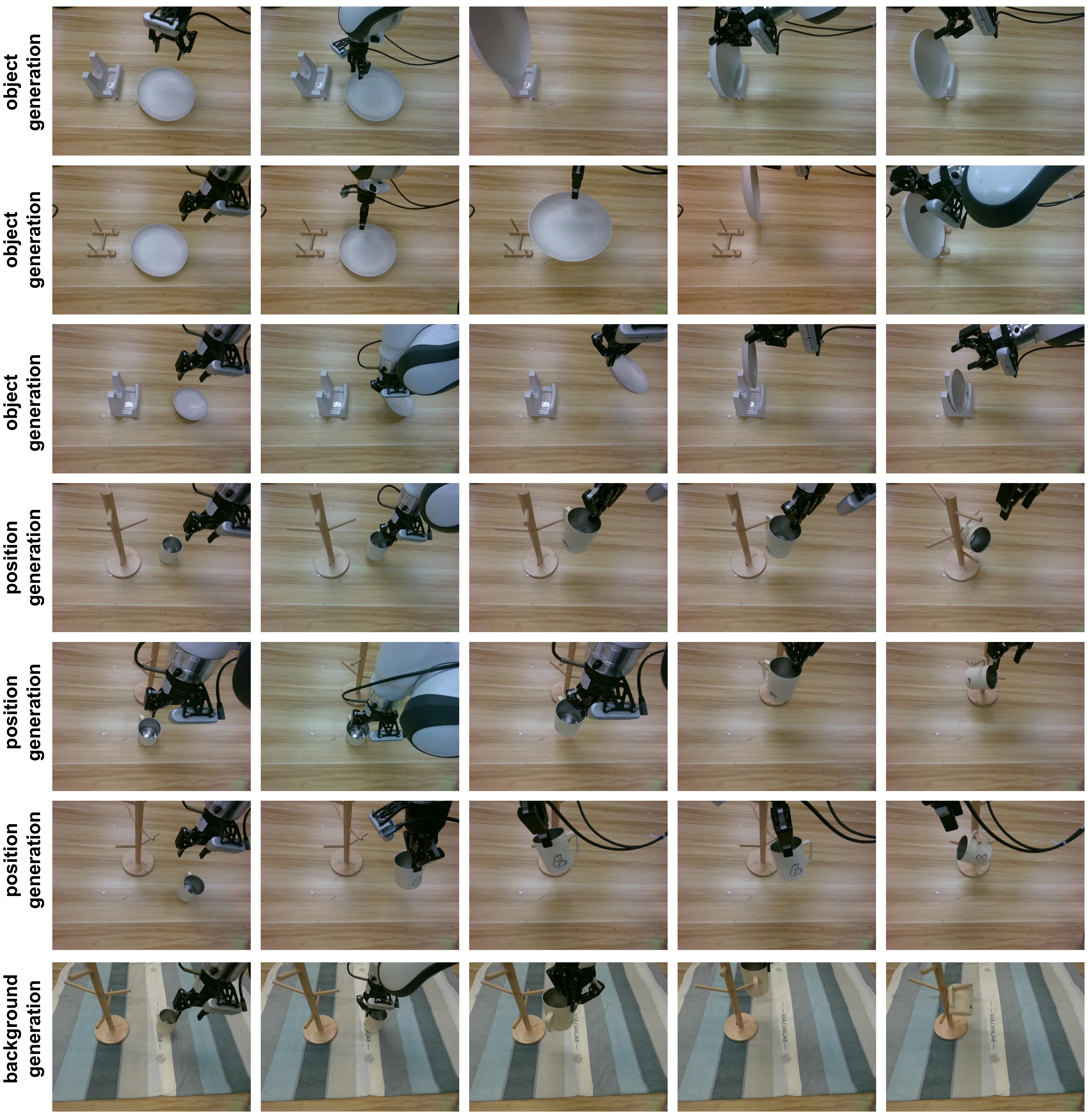}
  \caption{\textbf{Generalization of evolved skills in real-world manipulation.}
  \emph{Object generalization} (top three rows): the plate-placement
  skill is applied to different plate and rack configurations, including
  smaller objects. \emph{Position generalization} (middle three rows):
  the mug-hanging skill is reused from different initial mug positions.
  \emph{Background generalization} (bottom row): the same mug-hanging
  skill is executed with a changed visual background. These sequences
  illustrate adaptation beyond the original task configurations.}
  \label{fig:supp_skill_generalization}
\end{figure*}

\begin{figure*}[t]
    \centering
    \includegraphics[width=\textwidth]
        {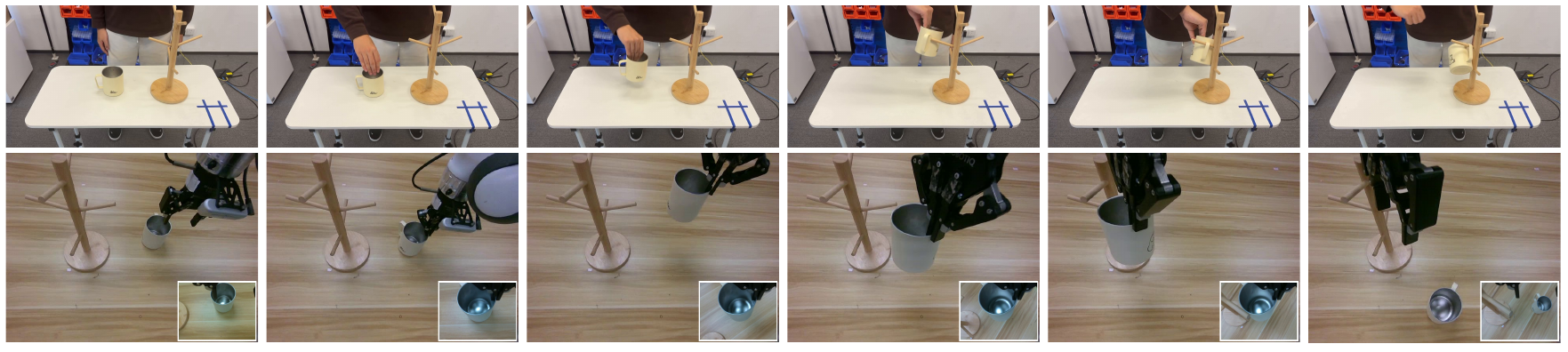}
    \caption{
        Direct conditioning on a human demonstration for hanging
        a mug.
        Top: successful human demonstration.
        Bottom: robot execution conditioned on the demonstration,
        without Real2Gym's iterative skill refinement.
        Lower-right insets show the corresponding wrist-camera views.
        The robot does not achieve the demonstrated hanging state,
        and the mug ends on the tabletop.
    }
    \label{fig:ablation_demonstration}
\end{figure*}

To examine whether simulation-acquired skills are reusable rather than
restricted to one fixed execution trajectory, we evaluate additional
real-world configurations for the two challenging tasks
(Fig.~\ref{fig:supp_skill_generalization}). For \emph{mug hanging}, we vary the initial mug position relative to the rack and separately change the scene background, while consistently keeping the mug handle occluded from the initial camera view. Position variation requires the agent to adapt its reaching, grasping, and alignment behavior, while background variation tests its robustness to changes in visual appearance. Importantly, the initially occluded handle introduces an additional manipulation challenge, requiring the agent to actively reorient the mug to expose the handle before aligning it with the rack. The successful sequences demonstrate that the acquired mug-hanging skill can be effectively reused across different spatial configurations and visual backgrounds, even when the task-critical handle is initially invisible.

For \emph{plate placement}, we test object generalization by changing
the plate and rack, including configurations with smaller objects.
The resulting geometric differences alter the grasp and placement
constraints, making a literal replay of the original motion
insufficient. As shown in Fig.~\ref{fig:supp_skill_generalization},
the agent adapts its manipulation to the changed object geometry and
successfully completes the placement. Together, these qualitative
results support the transferability of evolved skills across variations
in object geometry, initial spatial configuration, and visual context.

Overall, the supplementary sequences illustrate the full Real2Sim2Real
process: human demonstrations are used to establish interactive
simulation environments, simulation experience is distilled into
evolved skills, and these skills improve real-world execution while
remaining applicable to previously unseen task configurations.

\subsection{Direct conditioning on human demonstrations}

We provide the agent with a human demonstration as reference
context, without the iterative skill refinement used in Real2Gym.
Fig.~\ref{fig:ablation_demonstration} shows representative
keyframes from the demonstration and robot execution.
Although the human successfully hangs the mug, the agent does
not complete the task.
This illustrates that task-level guidance from human demonstrations
does not necessarily translate into actionable, embodiment-specific
skills for the robot.
Real2Gym refines skills through iterative interaction from the
robot's perspective, allowing observed execution failures to inform
subsequent corrections.

\section{Ablation on VLMs and Reasoning Effort}
\label{app:model_effort_ablation}

\begin{figure*}[t!]
    \centering
    \includegraphics[
        width=\linewidth,
        height=0.68\textheight,
        keepaspectratio
    ]{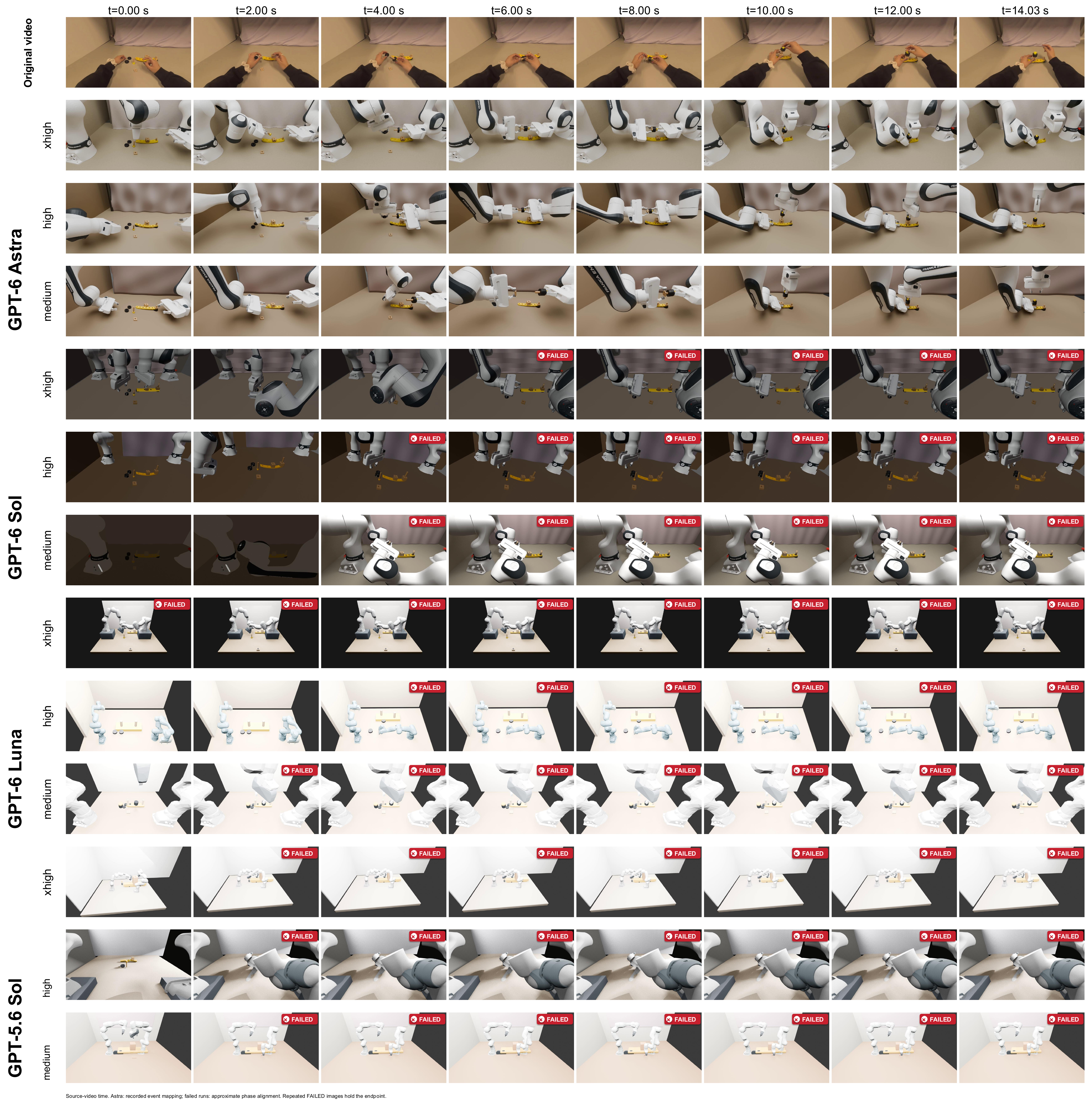}
    \caption{
        \textbf{Qualitative comparison across models and reasoning effort levels on an EgoDex assembly task.} The top row depicts the source human demonstration, while subsequent rows show the 12 experimental configurations. Columns are indexed by source-video timestamps. Successful trajectories are aligned using recorded event correspondences, whereas failed attempts are approximately aligned by the attempted action phase. Repeated frames marked with \texttt{FAILED} retain a frozen state for visual comparison and do not imply continued execution.
    }
    \label{fig:model_effort_ablation}
\end{figure*}

We evaluate the effects of model choice and reasoning effort on a challenging bimanual assembly task from EgoDex. The source demonstration involves placing two wheels and a square nut onto a bolt, tightening the nut, and inserting the assembly into a base. Reproducing this sequence requires stable grasping, precise hole--shaft alignment, and contact dynamics coupling rotation with axial motion. Fig.~\ref{fig:model_effort_ablation} compares four models across three reasoning effort levels (medium, high, and xhigh), yielding 12 configurations. GPT-6 Astra and GPT-6 Sol generally recover the main objects and scene layout in reasonable agreement with the source video, although Sol retains visual alignment and occlusion discrepancies. Only GPT-6 Astra passes the full-task simulation criteria at all three effort levels, completing an adapted execution that includes nut tightening and final insertion. GPT-6 Sol at high and medium effort fails during grasping or before assembling the first wheel, while xhigh reaches the second-wheel stage but fails to release it; none reaches successful nut tightening. GPT-6 Luna and GPT-5.6 Sol exhibit larger discrepancies in scene geometry, object relationships, or Viewpoint Alignment. Their final candidates fail during initialization, early grasping, or local manipulation checks and do not reach the full assembly sequence. This case study highlights the gap between visually plausible scene reconstruction and physically executable, long-horizon manipulation.

\FloatBarrier
\section{Per-Task Real2Sim Results}
\label{app:real2sim_results}

Tab.~\ref{tab:real2sim_per_task} through Tab.~\ref{tab:real2sim_per_task_ours}
report task-level Real2Sim results on DROID~\citep{droid} and
EgoDex~\citep{egodex} for GPT-5.6 Sol, GPT-6 Astra,
Agentic Real2Sim~\citep{chen2026agenticreal2sim}, and Real2Gym.
The four rubric-based metrics assess Content Alignment,
Viewpoint Alignment, Action Fidelity, and Simulation Success
(the observable completion subtotal, scored out of 80).
Each rubric entry lists scores from GPT-6 Astra, GPT-6.1 Sol,
and Claude Opus 5.5, in that order.
We additionally report LPIPS-VGG and DINOv2 patch cosine
similarity to measure initial-frame visual correspondence.
Dataset means are computed over valid scores, excluding missing
entries. Agentic Real2Sim results cover its ten available DROID
outputs and are not applicable to EgoDex.
The evaluation criteria are provided in
Sec.~\ref{app:evaluation_prompt}.
Overall, Real2Gym shows stronger visual alignment and action
fidelity across both datasets, although improvements vary by
task and judge.
\begin{table*}[!htbp]
\centering
\caption{Per-task Real2Sim evaluation on DROID and EgoDex. Each rubric entry reports GPT-6 Astra / GPT-6.1 Sol / Claude Opus 5.5, in that order. Content, viewpoint and action are out of 100; simulation success is the observable subtotal out of 80. LPIPS-VGG and DINOv2 patch cosine compare native initial frames. Bold indicates the best available score per task, metric and judge, including ties. Means use valid values only; no missing value is replaced by zero.}
\label{tab:real2sim_per_task}
\scriptsize
\setlength{\tabcolsep}{3pt}
\renewcommand{\arraystretch}{1.13}
\resizebox{\textwidth}{!}{%
\begin{tabular}{@{}l*{6}{c}@{}}
\toprule
\multicolumn{7}{c}{\textbf{GPT-5.6 Sol xhigh}} \\
\midrule
\textbf{Task} & \shortstack{Content\\Alignment $\uparrow$} & \shortstack{Viewpoint\\Alignment $\uparrow$} & \shortstack{Action\\Fidelity $\uparrow$} & \shortstack{Simulation\\Success (/80) $\uparrow$} & LPIPS-VGG $\downarrow$ & DINOv2 Cos. $\uparrow$ \\
\midrule
\multicolumn{7}{l}{\textbf{DROID}} \\
D01 & 52.00 / 50.00 / 47.00 & 30.00 / 35.00 / 45.00 & 84.00 / 77.00 / \textbf{78.00} & 56.00 / 67.00 / \textbf{80.00} & 0.8763 & 0.2053 \\
D02 & 47.00 / 53.00 / 52.00 & 32.00 / 35.00 / 30.00 & 72.00 / 77.00 / -- & 64.00 / 67.00 / -- & 0.7068 & 0.1773 \\
D03 & 43.00 / 40.00 / 40.00 & 33.00 / 35.00 / 15.00 & 77.00 / 80.00 / -- & 68.00 / 68.00 / -- & 0.7143 & 0.1798 \\
D04 & 50.00 / 48.00 / 50.00 & 45.00 / 40.00 / 45.00 & 63.00 / 60.00 / 68.00 & 46.00 / 48.00 / 50.00 & 0.8003 & 0.1491 \\
D05 & 55.00 / 52.00 / 56.00 & 30.00 / 35.00 / 25.00 & \textbf{40.00} / 75.00 / 30.00 & 60.00 / 66.00 / 38.00 & 0.6683 & 0.2289 \\
D06 & 48.00 / 51.00 / 47.00 & 20.00 / 25.00 / 15.00 & 78.00 / 81.00 / 71.00 & 76.00 / 77.00 / \textbf{75.00} & 0.7439 & 0.1335 \\
D07 & 56.00 / 54.00 / 60.00 & 34.00 / 34.00 / 35.00 & 74.00 / 80.00 / \textbf{75.00} & 68.00 / 73.00 / \textbf{72.00} & 0.6792 & 0.1709 \\
D08 & 27.00 / 22.00 / 33.00 & 20.00 / 23.00 / 20.00 & 72.00 / 76.00 / 76.00 & 56.00 / 63.00 / 64.00 & 0.7154 & 0.1372 \\
D09 & 52.00 / 58.00 / 59.00 & 34.00 / 40.00 / 25.00 & 82.00 / 80.00 / -- & 72.00 / 70.00 / -- & 0.7503 & 0.1860 \\
D10 & 49.00 / 52.00 / 50.00 & 36.00 / 42.00 / 30.00 & 67.00 / 74.00 / 69.00 & 58.00 / 62.00 / 50.00 & 0.6688 & 0.2123 \\
D11 & 29.00 / 29.00 / 30.00 & 25.00 / 25.00 / 25.00 & 57.00 / 34.00 / 35.00 & 34.00 / 36.00 / 30.00 & 0.8446 & 0.1580 \\
D12 & 45.00 / 48.00 / 54.00 & 25.00 / 30.00 / 20.00 & 64.00 / 68.00 / 71.00 & 52.00 / 47.00 / 59.00 & 0.7169 & 0.1440 \\
\midrule
\textbf{Mean} & 46.08 / 46.42 / 48.17 & 30.33 / 33.25 / 27.50 & 69.17 / 71.83 / 63.67 & 59.17 / 62.00 / 57.56 & 0.7404 & 0.1735 \\
\midrule
\multicolumn{7}{l}{\textbf{EgoDex}} \\
E01 & 67.00 / 68.00 / 71.00 & 55.00 / 58.00 / 50.00 & 87.00 / 88.00 / \textbf{88.00} & 77.00 / 77.00 / \textbf{75.00} & 0.6771 & 0.1881 \\
E02 & 55.00 / 53.00 / 56.00 & 50.00 / 42.00 / 55.00 & 83.00 / 83.00 / 79.00 & 76.00 / 75.00 / 68.00 & 0.6514 & 0.1636 \\
E03 & 69.00 / 72.00 / 68.00 & 30.00 / 32.00 / 45.00 & 76.00 / 80.00 / 65.00 & 50.00 / 54.00 / 57.00 & 0.6682 & 0.1679 \\
E04 & 75.00 / 68.00 / 71.00 & 67.00 / 64.00 / 75.00 & 83.00 / 84.00 / 87.00 & 70.00 / 73.00 / \textbf{73.00} & 0.6423 & 0.1773 \\
E05 & 63.00 / 66.00 / 60.00 & 30.00 / 30.00 / 25.00 & 65.00 / 66.00 / 57.00 & 55.00 / 62.00 / 57.00 & 0.7195 & 0.1366 \\
E06 & 67.00 / 69.00 / 61.00 & 29.00 / 30.00 / 30.00 & 82.00 / 83.00 / -- & 74.00 / 74.00 / \textbf{80.00} & 0.6729 & 0.1579 \\
E07 & 63.00 / 63.00 / 63.00 & 35.00 / 42.00 / 30.00 & 71.00 / 78.00 / -- & 68.00 / 70.00 / 67.00 & 0.7342 & 0.1733 \\
E08 & 66.00 / 65.00 / 56.00 & \textbf{44.00} / 52.00 / 45.00 & 77.00 / 80.00 / 64.00 & 64.00 / 74.00 / 52.00 & 0.6962 & 0.1851 \\
E09 & 56.00 / 54.00 / 52.00 & 46.00 / 43.00 / 35.00 & 64.00 / 61.00 / 49.00 & 42.00 / 40.00 / 37.00 & 0.6474 & 0.1370 \\
E10 & 57.00 / 60.00 / 63.00 & 38.00 / 42.00 / 40.00 & 70.00 / 66.00 / 72.00 & 62.00 / 52.00 / 71.00 & 0.8263 & 0.1389 \\
E11 & 54.00 / 63.00 / 62.00 & 29.00 / 35.00 / 45.00 & 63.00 / 66.00 / -- & 57.00 / 62.00 / 66.00 & 0.7746 & 0.1213 \\
E12 & 48.00 / 52.00 / 42.00 & 44.00 / 52.00 / 45.00 & 32.00 / 32.00 / 26.00 & 16.00 / 17.00 / 9.00 & 0.6541 & 0.1764 \\
\midrule
\textbf{Mean} & 61.67 / 62.75 / 60.42 & 41.42 / 43.50 / 43.33 & 71.08 / 72.25 / 65.22 & 59.25 / 60.83 / 59.33 & 0.6970 & 0.1603 \\
\bottomrule
\end{tabular}%
}
\vspace{2pt}
\begin{minipage}{\textwidth}\scriptsize
Judges: GPT-6 Astra / GPT-6.1 Sol / Claude Opus 5.5. $--$: indicates that the reconstruction quality was too low for the judge to apply the scoring rubric; Fail: no native output; N/A: not applicable. Means use valid scores only; coverage differs across methods.
\end{minipage}
\end{table*}

\begin{table*}[!htbp]
\centering
\caption{Per-task Real2Sim evaluation: GPT-6 Astra Medium. Judge order and scoring scales follow Tab.~\ref{tab:real2sim_per_task}.}
\label{tab:real2sim_per_task_astra}
\scriptsize
\setlength{\tabcolsep}{3pt}
\renewcommand{\arraystretch}{1.13}
\resizebox{\textwidth}{!}{%
\begin{tabular}{@{}l*{6}{c}@{}}
\toprule
\multicolumn{7}{c}{\textbf{GPT-6 Astra Medium}} \\
\midrule
\textbf{Task} & \shortstack{Content\\Alignment $\uparrow$} & \shortstack{Viewpoint\\Alignment $\uparrow$} & \shortstack{Action\\Fidelity $\uparrow$} & \shortstack{Simulation\\Success (/80) $\uparrow$} & LPIPS-VGG $\downarrow$ & DINOv2 Cos. $\uparrow$ \\
\midrule
\multicolumn{7}{l}{\textbf{DROID}} \\
D01 & 40.00 / 62.00 / 63.00 & 28.00 / 38.00 / 30.00 & 78.00 / 87.00 / 71.00 & \textbf{76.00} / 74.00 / 60.00 & 0.8523 & 0.2363 \\
D02 & 55.00 / \textbf{73.00} / 66.00 & 32.00 / 36.00 / 25.00 & 82.00 / 84.00 / 59.00 & 73.00 / 73.00 / 44.00 & \textbf{0.6677} & 0.2423 \\
D03 & 51.00 / \textbf{65.00} / 68.00 & 39.00 / 48.00 / 50.00 & 83.00 / 86.00 / 79.00 & 71.00 / 73.00 / 65.00 & 0.7433 & 0.2172 \\
D04 & 45.00 / \textbf{70.00} / 64.00 & 35.00 / 40.00 / 30.00 & 62.00 / 72.00 / 66.00 & 49.00 / 55.00 / 48.00 & 0.7641 & 0.2003 \\
D05 & 51.00 / \textbf{70.00} / 64.00 & 30.00 / 35.00 / 30.00 & \textbf{40.00} / \textbf{87.00} / 57.00 & 40.00 / \textbf{75.00} / 47.00 & 0.7183 & 0.2488 \\
D06 & 55.00 / 56.00 / 59.00 & 25.00 / 25.00 / 25.00 & 88.00 / 88.00 / 80.00 & 75.00 / 78.00 / 71.00 & 0.7129 & 0.2211 \\
D07 & 58.00 / 66.00 / 62.00 & 29.00 / 42.00 / 30.00 & 85.00 / 86.00 / 54.00 & 78.00 / \textbf{74.00} / 42.00 & 0.6891 & 0.2000 \\
D08 & 35.00 / 49.00 / 56.00 & 38.00 / 40.00 / 50.00 & 85.00 / 87.00 / 76.00 & 76.00 / \textbf{80.00} / 66.00 & 0.7073 & 0.2023 \\
D09 & 51.00 / 57.00 / 58.00 & 35.00 / 50.00 / 30.00 & 75.00 / 75.00 / 61.00 & 71.00 / 67.00 / 60.00 & 0.7739 & 0.2009 \\
D10 & 55.00 / 70.00 / 60.00 & 30.00 / 32.00 / 28.00 & 82.00 / 82.00 / 66.00 & \textbf{74.00} / \textbf{74.00} / 63.00 & 0.7400 & 0.1883 \\
D11 & 32.00 / 36.00 / 36.00 & 20.00 / 24.00 / 22.00 & 64.00 / 56.00 / 47.00 & 67.00 / 54.00 / 42.00 & 0.7912 & 0.1297 \\
D12 & 40.00 / 49.00 / 56.00 & 25.00 / 28.00 / 30.00 & 70.00 / 76.00 / 71.00 & 60.00 / 59.00 / 63.00 & 0.7202 & 0.1747 \\
\midrule
\textbf{Mean} & 47.33 / 60.25 / 59.33 & 30.50 / 36.50 / 31.67 & 74.50 / 80.50 / 65.58 & 67.50 / 69.67 / 55.92 & 0.7400 & 0.2051 \\
\midrule
\multicolumn{7}{l}{\textbf{EgoDex}} \\
E01 & 61.00 / 65.00 / 62.00 & 36.00 / 38.00 / 40.00 & 86.00 / 87.00 / 79.00 & \textbf{80.00} / \textbf{80.00} / 69.00 & 0.7205 & 0.1869 \\
E02 & 73.00 / 65.00 / 70.00 & 44.00 / 45.00 / 38.00 & 87.00 / \textbf{90.00} / 74.00 & \textbf{80.00} / \textbf{80.00} / 70.00 & 0.6666 & 0.1936 \\
E03 & 70.00 / 70.00 / 66.00 & 46.00 / 45.00 / 58.00 & 63.00 / 56.00 / 65.00 & 36.00 / 34.00 / 55.00 & 0.6806 & \textbf{0.1870} \\
E04 & 68.00 / 64.00 / 66.00 & 40.00 / 43.00 / 45.00 & 83.00 / 84.00 / 81.00 & 71.00 / 71.00 / 66.00 & 0.7430 & 0.1946 \\
E05 & 57.00 / 52.00 / 59.00 & 32.00 / 30.00 / 40.00 & 75.00 / 86.00 / 59.00 & 78.00 / 71.00 / 52.00 & 0.6779 & 0.1564 \\
E06 & 63.00 / 66.00 / 68.00 & 30.00 / 32.00 / 20.00 & 84.00 / 88.00 / 82.00 & 78.00 / \textbf{80.00} / 74.00 & 0.6850 & 0.1660 \\
E07 & 68.00 / 64.00 / 71.00 & 43.00 / 52.00 / 60.00 & 68.00 / 66.00 / 66.00 & 47.00 / 46.00 / 54.00 & 0.7872 & \textbf{0.1982} \\
E08 & 61.00 / 66.00 / \textbf{70.00} & 43.00 / \textbf{54.00} / \textbf{75.00} & 60.00 / 59.00 / 54.00 & 37.00 / 37.00 / 32.00 & 0.6811 & \textbf{0.2044} \\
E09 & 54.00 / 56.00 / 60.00 & 32.00 / 42.00 / 60.00 & 64.00 / 67.00 / 60.00 & 42.00 / 46.00 / 40.00 & 0.6696 & 0.1723 \\
E10 & 61.00 / 61.00 / 60.00 & 28.00 / 28.00 / 35.00 & 73.00 / 72.00 / 66.00 & 66.00 / 63.00 / 58.00 & 0.7305 & 0.1489 \\
E11 & 62.00 / 59.00 / 60.00 & 29.00 / 32.00 / 40.00 & 78.00 / 86.00 / 81.00 & 67.00 / 68.00 / 67.00 & 0.7256 & 0.1554 \\
E12 & 60.00 / 66.00 / 55.00 & 34.00 / 42.00 / 32.00 & 40.00 / 40.00 / 40.00 & 50.00 / 52.00 / 42.00 & 0.6705 & 0.2027 \\
\midrule
\textbf{Mean} & 63.17 / 62.83 / 63.92 & 36.42 / 40.25 / 45.25 & 71.75 / 73.42 / 67.25 & 61.00 / 60.67 / 56.58 & 0.7032 & 0.1805 \\
\bottomrule
\end{tabular}%
}
\vspace{2pt}
\begin{minipage}{\textwidth}\scriptsize
Judges: GPT-6 Astra / GPT-6.1 Sol / Claude Opus 5.5. $--$: indicates that the reconstruction quality was too low for the judge to apply the scoring rubric; Fail: no native output; N/A: not applicable. Means use valid scores only; coverage differs across methods.
\end{minipage}
\end{table*}

\begin{table*}[!htbp]
\centering
\caption{Per-task Real2Sim evaluation: Agentic Real2Sim. Judge order and scoring scales follow Tab.~\ref{tab:real2sim_per_task}.}
\label{tab:real2sim_per_task_agentic}
\scriptsize
\setlength{\tabcolsep}{3pt}
\renewcommand{\arraystretch}{1.13}
\resizebox{\textwidth}{!}{%
\begin{tabular}{@{}l*{6}{c}@{}}
\toprule
\multicolumn{7}{c}{\textbf{Agentic Real2Sim}} \\
\midrule
\textbf{Task} & \shortstack{Content\\Alignment $\uparrow$} & \shortstack{Viewpoint\\Alignment $\uparrow$} & \shortstack{Action\\Fidelity $\uparrow$} & \shortstack{Simulation\\Success (/80) $\uparrow$} & LPIPS-VGG $\downarrow$ & DINOv2 Cos. $\uparrow$ \\
\midrule
\multicolumn{7}{l}{\textbf{DROID}} \\
D01 & 59.00 / 59.00 / 60.00 & 35.00 / 35.00 / 30.00 & 40.00 / 40.00 / 40.00 & 34.00 / -- / 20.00 & 0.8332 & 0.1970 \\
D02 & 51.00 / 52.00 / 61.00 & 38.00 / 35.00 / 45.00 & 40.00 / 70.00 / 40.00 & 44.00 / 60.00 / 32.00 & 0.7123 & 0.1798 \\
D03 & 42.00 / 37.00 / 49.00 & 25.00 / 25.00 / 25.00 & 38.00 / 46.00 / 40.00 & 34.00 / 32.00 / 26.00 & 0.7460 & 0.1368 \\
D04 & 29.00 / 27.00 / 30.00 & 28.00 / 35.00 / 30.00 & 40.00 / 40.00 / 40.00 & 50.00 / 60.00 / 54.00 & 0.7434 & 0.1485 \\
D05 & Fail & Fail & Fail & Fail & Fail & Fail \\
D06 & 39.00 / 46.00 / 40.00 & 48.00 / 40.00 / 40.00 & 27.00 / 33.00 / 25.00 & 15.00 / 15.00 / 8.00 & 0.7177 & 0.1734 \\
D07 & 24.00 / 28.00 / 28.00 & 8.00 / 5.00 / 10.00 & 13.00 / 15.00 / 28.00 & 5.00 / 10.00 / 14.00 & 0.7715 & 0.1248 \\
D08 & 40.00 / 42.00 / 30.00 & 25.00 / 30.00 / 30.00 & 0.00 / 0.00 / 0.00 & 0.00 / 0.00 / 0.00 & 0.7055 & 0.1750 \\
D09 & 22.00 / 26.00 / 24.00 & 25.00 / 30.00 / 30.00 & 25.00 / 23.00 / 26.00 & 8.00 / 10.00 / 8.00 & 0.7779 & 0.1511 \\
D10 & 55.00 / 59.00 / 56.00 & 25.00 / 28.00 / 25.00 & 74.00 / 77.00 / 85.00 & 69.00 / 67.00 / -- & 0.7109 & 0.1520 \\
D11 & Fail & Fail & Fail & Fail & Fail & Fail \\
D12 & 39.00 / 36.00 / 30.00 & 32.00 / 32.00 / 45.00 & 64.00 / 63.00 / 83.00 & 37.00 / 40.00 / 60.00 & 0.6994 & 0.1256 \\
\midrule
\textbf{Mean} & 40.00 / 41.20 / 40.80 & 28.90 / 29.50 / 31.00 & 36.10 / 40.70 / 40.70 & 29.60 / 32.67 / 24.67 & 0.7418 & 0.1564 \\
\midrule
\multicolumn{7}{l}{\textbf{EgoDex}} \\
E01 & N/A & N/A & N/A & N/A & N/A & N/A \\
E02 & N/A & N/A & N/A & N/A & N/A & N/A \\
E03 & N/A & N/A & N/A & N/A & N/A & N/A \\
E04 & N/A & N/A & N/A & N/A & N/A & N/A \\
E05 & N/A & N/A & N/A & N/A & N/A & N/A \\
E06 & N/A & N/A & N/A & N/A & N/A & N/A \\
E07 & N/A & N/A & N/A & N/A & N/A & N/A \\
E08 & N/A & N/A & N/A & N/A & N/A & N/A \\
E09 & N/A & N/A & N/A & N/A & N/A & N/A \\
E10 & N/A & N/A & N/A & N/A & N/A & N/A \\
E11 & N/A & N/A & N/A & N/A & N/A & N/A \\
E12 & N/A & N/A & N/A & N/A & N/A & N/A \\
\midrule
\textbf{Mean} & N/A & N/A & N/A & N/A & N/A & N/A \\
\bottomrule
\end{tabular}%
}
\vspace{2pt}
\begin{minipage}{\textwidth}\scriptsize
Judges: GPT-6 Astra / GPT-6.1 Sol / Claude Opus 5.5. $--$: indicates that the reconstruction quality was too low for the judge to apply the scoring rubric; Fail: no native output; N/A: not applicable. Means use valid scores only; coverage differs across methods.
\end{minipage}
\end{table*}

\begin{table*}[!htbp]
\centering
\caption{Per-task Real2Sim evaluation: Real2Gym (Ours). Judge order and scoring scales follow Tab.~\ref{tab:real2sim_per_task}.}
\label{tab:real2sim_per_task_ours}
\scriptsize
\setlength{\tabcolsep}{3pt}
\renewcommand{\arraystretch}{1.13}
\resizebox{\textwidth}{!}{%
\begin{tabular}{@{}l*{6}{c}@{}}
\toprule
\multicolumn{7}{c}{\textbf{Real2Gym (Ours)}} \\
\midrule
\textbf{Task} & \shortstack{Content\\Alignment $\uparrow$} & \shortstack{Viewpoint\\Alignment $\uparrow$} & \shortstack{Action\\Fidelity $\uparrow$} & \shortstack{Simulation\\Success (/80) $\uparrow$} & LPIPS-VGG $\downarrow$ & DINOv2 Cos. $\uparrow$ \\
\midrule
\multicolumn{7}{l}{\textbf{DROID}} \\
D01 & \textbf{90.00} / \textbf{73.00} / \textbf{66.00} & \textbf{90.00} / \textbf{94.00} / \textbf{80.00} & \textbf{95.00} / \textbf{96.00} / 75.00 & \textbf{76.00} / \textbf{78.00} / 60.00 & \textbf{0.5130} & \textbf{0.4230} \\
D02 & \textbf{85.00} / 68.00 / \textbf{74.00} & \textbf{86.00} / \textbf{91.00} / \textbf{85.00} & \textbf{93.00} / \textbf{90.00} / \textbf{80.00} & \textbf{77.00} / \textbf{75.00} / \textbf{69.00} & 0.7397 & \textbf{0.3740} \\
D03 & \textbf{79.00} / \textbf{65.00} / \textbf{76.00} & \textbf{84.00} / \textbf{84.00} / \textbf{88.00} & \textbf{95.00} / \textbf{95.00} / \textbf{87.00} & \textbf{78.00} / \textbf{80.00} / \textbf{75.00} & \textbf{0.6735} & \textbf{0.2867} \\
D04 & \textbf{83.00} / 65.00 / \textbf{75.00} & \textbf{80.00} / \textbf{88.00} / \textbf{85.00} & \textbf{93.00} / \textbf{92.00} / \textbf{87.00} & \textbf{77.00} / \textbf{77.00} / \textbf{72.00} & \textbf{0.7010} & \textbf{0.3472} \\
D05 & \textbf{84.00} / 64.00 / \textbf{68.00} & \textbf{85.00} / \textbf{90.00} / \textbf{76.00} & \textbf{40.00} / 64.00 / \textbf{81.00} & \textbf{68.00} / 44.00 / \textbf{67.00} & \textbf{0.5497} & \textbf{0.3639} \\
D06 & \textbf{77.00} / \textbf{71.00} / \textbf{72.00} & \textbf{82.00} / \textbf{91.00} / \textbf{85.00} & \textbf{97.00} / \textbf{96.00} / \textbf{93.00} & \textbf{78.00} / \textbf{80.00} / \textbf{75.00} & \textbf{0.5537} & \textbf{0.3584} \\
D07 & \textbf{84.00} / \textbf{73.00} / \textbf{76.00} & \textbf{87.00} / \textbf{88.00} / \textbf{60.00} & \textbf{96.00} / \textbf{92.00} / \textbf{75.00} & \textbf{79.00} / \textbf{74.00} / 54.00 & \textbf{0.5668} & \textbf{0.4840} \\
D08 & \textbf{78.00} / \textbf{73.00} / \textbf{73.00} & \textbf{82.00} / \textbf{84.00} / \textbf{80.00} & \textbf{96.00} / \textbf{95.00} / \textbf{85.00} & \textbf{80.00} / \textbf{80.00} / \textbf{72.00} & \textbf{0.6668} & \textbf{0.2973} \\
D09 & \textbf{77.00} / \textbf{71.00} / \textbf{70.00} & \textbf{82.00} / \textbf{88.00} / \textbf{80.00} & \textbf{95.00} / \textbf{92.00} / \textbf{79.00} & \textbf{79.00} / \textbf{80.00} / \textbf{66.00} & \textbf{0.7076} & \textbf{0.3152} \\
D10 & \textbf{86.00} / \textbf{74.00} / \textbf{74.00} & \textbf{77.00} / \textbf{88.00} / \textbf{80.00} & \textbf{92.00} / \textbf{89.00} / \textbf{88.00} & 72.00 / 68.00 / \textbf{68.00} & \textbf{0.5044} & \textbf{0.3854} \\
D11 & \textbf{78.00} / \textbf{64.00} / \textbf{82.00} & \textbf{84.00} / \textbf{84.00} / \textbf{85.00} & \textbf{89.00} / \textbf{88.00} / \textbf{78.00} & \textbf{74.00} / \textbf{70.00} / \textbf{67.00} & \textbf{0.6031} & \textbf{0.4095} \\
D12 & \textbf{74.00} / \textbf{68.00} / \textbf{74.00} & \textbf{73.00} / \textbf{70.00} / \textbf{85.00} & \textbf{93.00} / \textbf{90.00} / \textbf{87.00} & \textbf{77.00} / \textbf{73.00} / \textbf{69.00} & \textbf{0.6254} & \textbf{0.3175} \\
\midrule
\textbf{Mean} & 81.25 / 69.08 / 73.33 & 82.67 / 86.67 / 80.75 & 89.50 / 89.92 / 82.92 & 76.25 / 73.25 / 67.83 & 0.6171 & 0.3635 \\
\midrule
\multicolumn{7}{l}{\textbf{EgoDex}} \\
E01 & \textbf{82.00} / \textbf{79.00} / \textbf{75.00} & \textbf{90.00} / \textbf{94.00} / \textbf{78.00} & \textbf{95.00} / \textbf{94.00} / 85.00 & \textbf{80.00} / \textbf{80.00} / 68.00 & \textbf{0.5293} & \textbf{0.2364} \\
E02 & \textbf{80.00} / \textbf{76.00} / \textbf{78.00} & \textbf{76.00} / \textbf{86.00} / \textbf{62.00} & \textbf{92.00} / \textbf{90.00} / \textbf{85.00} & \textbf{80.00} / \textbf{80.00} / \textbf{72.00} & \textbf{0.5296} & \textbf{0.2859} \\
E03 & \textbf{82.00} / \textbf{78.00} / \textbf{77.00} & \textbf{77.00} / \textbf{86.00} / \textbf{78.00} & \textbf{94.00} / \textbf{92.00} / \textbf{84.00} & \textbf{79.00} / \textbf{76.00} / \textbf{66.00} & \textbf{0.5333} & 0.1722 \\
E04 & \textbf{83.00} / \textbf{80.00} / \textbf{81.00} & \textbf{78.00} / \textbf{93.00} / \textbf{85.00} & \textbf{96.00} / \textbf{96.00} / \textbf{92.00} & \textbf{80.00} / \textbf{80.00} / \textbf{73.00} & \textbf{0.5591} & \textbf{0.2457} \\
E05 & \textbf{85.00} / \textbf{89.00} / \textbf{86.00} & \textbf{80.00} / \textbf{92.00} / \textbf{80.00} & \textbf{93.00} / \textbf{96.00} / \textbf{86.00} & \textbf{80.00} / \textbf{80.00} / \textbf{71.00} & \textbf{0.5797} & \textbf{0.1770} \\
E06 & \textbf{84.00} / \textbf{84.00} / \textbf{78.00} & \textbf{85.00} / \textbf{92.00} / \textbf{85.00} & \textbf{93.00} / \textbf{95.00} / \textbf{86.00} & \textbf{80.00} / \textbf{80.00} / 74.00 & \textbf{0.5392} & \textbf{0.2907} \\
E07 & \textbf{82.00} / \textbf{76.00} / \textbf{78.00} & \textbf{74.00} / \textbf{90.00} / \textbf{72.00} & \textbf{87.00} / \textbf{85.00} / \textbf{84.00} & \textbf{76.00} / \textbf{74.00} / \textbf{71.00} & \textbf{0.7014} & 0.1867 \\
E08 & \textbf{73.00} / \textbf{72.00} / 69.00 & 27.00 / 32.00 / 45.00 & \textbf{89.00} / \textbf{84.00} / \textbf{81.00} & \textbf{77.00} / \textbf{76.00} / \textbf{67.00} & \textbf{0.6174} & 0.2042 \\
E09 & \textbf{84.00} / \textbf{80.00} / \textbf{76.00} & \textbf{65.00} / \textbf{84.00} / \textbf{75.00} & \textbf{91.00} / \textbf{87.00} / \textbf{85.00} & \textbf{74.00} / \textbf{72.00} / \textbf{67.00} & \textbf{0.6268} & \textbf{0.2016} \\
E10 & \textbf{79.00} / \textbf{76.00} / \textbf{84.00} & \textbf{78.00} / \textbf{85.00} / \textbf{75.00} & \textbf{94.00} / \textbf{93.00} / \textbf{88.00} & \textbf{80.00} / \textbf{80.00} / \textbf{72.00} & \textbf{0.6388} & \textbf{0.1760} \\
E11 & \textbf{86.00} / \textbf{80.00} / \textbf{81.00} & \textbf{82.00} / \textbf{93.00} / \textbf{84.00} & \textbf{91.00} / \textbf{95.00} / \textbf{85.00} & \textbf{75.00} / \textbf{78.00} / \textbf{69.00} & \textbf{0.6297} & \textbf{0.2214} \\
E12 & \textbf{89.00} / \textbf{89.00} / \textbf{90.00} & \textbf{71.00} / \textbf{90.00} / \textbf{92.00} & \textbf{84.00} / \textbf{84.00} / \textbf{84.00} & \textbf{75.00} / \textbf{72.00} / \textbf{71.00} & \textbf{0.4803} & \textbf{0.2896} \\
\midrule
\textbf{Mean} & 82.42 / 79.92 / 79.42 & 73.58 / 84.75 / 75.92 & 91.58 / 90.92 / 85.42 & 78.00 / 77.33 / 70.08 & 0.5804 & 0.2239 \\
\bottomrule
\end{tabular}%
}
\vspace{2pt}
\begin{minipage}{\textwidth}\scriptsize
\end{minipage}
\end{table*}

\FloatBarrier

\begin{table*}[p]
\begin{minipage}{\textwidth}
\section{Per-Task Simulation Control Results}
\label{app:simulation_results}
Tab.~\ref{tab:per_task_results_by_method} compares task success
and execution efficiency across baselines and our method
with and without skills on DROID and EgoDex.
The baselines include $\pi_{0.5}$-DROID~\citep{intelligence2025pi_} and Agent as Policy~\citep{jia2026agent}.
With skills, our agent additionally solves D10, D11, and E11 and reduces token usage on most tasks, while several failures remain. Task-specific success checks are also central to manipulation benchmarks such as robosuite, RLBench, and LIBERO~\citep{r2g_zhu2020robosuite,r2g_james2019rlbench,liu2023libero}.
\end{minipage}
\par\medskip
\centering
\caption{\textbf{Per-task efficiency and success results} on DROID and EgoDex. Responses denote model responses, and tokens are reported in millions. Lower is better for response count, token usage, and time.}
\label{tab:per_task_results_by_method}
\setlength{\tabcolsep}{7pt}
\renewcommand{\arraystretch}{0.86}
\resizebox{0.92\textwidth}{!}{%
\begin{tabular}{lcccc@{\hspace{16pt}}lcccc}
\toprule

\multicolumn{10}{c}{\textbf{$\pi_{0.5}$-DROID}} \\
\cmidrule(lr){1-10}
& \multicolumn{4}{c}{\textbf{DROID}} & & \multicolumn{4}{c}{\textbf{EgoDex}} \\
\cmidrule(lr){2-5} \cmidrule(lr){7-10}
\textbf{Task} & \textbf{Success} & \textbf{Resp.}$\downarrow$ & \textbf{Tokens (M)}$\downarrow$ & \textbf{Time (min)}$\downarrow$
& \textbf{Task} & \textbf{Success} & \textbf{Resp.}$\downarrow$ & \textbf{Tokens (M)}$\downarrow$ & \textbf{Time (min)}$\downarrow$ \\
D1  & $\times$ & N/A & N/A & N/A & E1  & $\times$ & N/A & N/A & N/A \\
D2  & $\times$ & N/A & N/A & N/A & E2  & $\times$ & N/A & N/A & N/A \\
D3  & $\times$ & N/A & N/A & N/A & E3  & $\times$ & N/A & N/A & N/A \\
D4  & $\times$ & N/A & N/A & N/A & E4  & $\times$ & N/A & N/A & N/A \\
D5  & $\times$ & N/A & N/A & N/A & E5  & $\times$ & N/A & N/A & N/A \\
D6  & $\times$ & N/A & N/A & N/A & E6  & $\times$ & N/A & N/A & N/A \\
D7  & $\times$ & N/A & N/A & N/A & E7  & $\times$ & N/A & N/A & N/A \\
D8  & $\times$ & N/A & N/A & N/A & E8  & $\times$ & N/A & N/A & N/A \\
D9  & $\times$ & N/A & N/A & N/A & E9  & $\times$ & N/A & N/A & N/A \\
D10 & $\times$ & N/A & N/A & N/A & E10 & $\times$ & N/A & N/A & N/A \\
D11 & $\times$ & N/A & N/A & N/A & E11 & $\times$ & N/A & N/A & N/A \\
D12 & $\times$ & N/A & N/A & N/A & E12 & $\times$ & N/A & N/A & N/A \\

\midrule
\multicolumn{10}{c}{\textbf{GPT-5.6 Sol xhigh}} \\
\cmidrule(lr){1-10}
& \multicolumn{4}{c}{\textbf{DROID}} & & \multicolumn{4}{c}{\textbf{EgoDex}} \\
\cmidrule(lr){2-5} \cmidrule(lr){7-10}
\textbf{Task} & \textbf{Success} & \textbf{Resp.}$\downarrow$ & \textbf{Tokens (M)}$\downarrow$ & \textbf{Time (min)}$\downarrow$
& \textbf{Task} & \textbf{Success} & \textbf{Resp.}$\downarrow$ & \textbf{Tokens (M)}$\downarrow$ & \textbf{Time (min)}$\downarrow$ \\
D1  & $\checkmark$ & 52  & 6.88  & 25.09 & E1  & $\checkmark$ & 75  & 9.12  & 36.88 \\
D2  & $\checkmark$ & 60  & 6.25  & 27.78 & E2  & $\checkmark$ & 59  & 6.12  & 14.76 \\
D3  & $\checkmark$ & 81  & 9.32  & 29.77 & E3  & $\times$     & 62  & 8.25  & 13.15 \\
D4  & $\times$     & 31  & 4.10  & 11.60 & E4  & $\checkmark$ & 42  & 4.27  & 15.44 \\
D5  & $\times$     & 53  & 6.44  & 28.45 & E5  & $\checkmark$ & 53  & 5.93  & 23.16 \\
D6  & $\checkmark$ & 56  & 4.83  & 23.43 & E6  & $\times$     & 81  & 12.50 & 24.11 \\
D7  & $\checkmark$ & 50  & 4.63  & 25.45 & E7  & $\times$     & 82  & 10.31 & 25.19 \\
D8  & $\checkmark$ & 62  & 6.69  & 29.81 & E8  & $\times$     & 317 & 41.32 & 65.55 \\
D9  & $\times$     & 39  & 5.53  & 16.91 & E9  & $\checkmark$ & 92  & 10.00 & 24.86 \\
D10 & $\times$     & 76  & 10.26 & 16.55 & E10 & $\times$     & 55  & 7.22  & 16.64 \\
D11 & $\times$     & 185 & 24.53 & 38.58 & E11 & $\times$     & 133 & 18.77 & 28.91 \\
D12 & $\checkmark$ & 112 & 13.37 & 35.22 & E12 & $\times$     & 215 & 29.77 & 56.05 \\

\midrule
\multicolumn{10}{c}{\textbf{GPT-6 Astra Medium}} \\
\cmidrule(lr){1-10}
& \multicolumn{4}{c}{\textbf{DROID}} & & \multicolumn{4}{c}{\textbf{EgoDex}} \\
\cmidrule(lr){2-5} \cmidrule(lr){7-10}
\textbf{Task} & \textbf{Success} & \textbf{Resp.}$\downarrow$ & \textbf{Tokens (M)}$\downarrow$ & \textbf{Time (min)}$\downarrow$
& \textbf{Task} & \textbf{Success} & \textbf{Resp.}$\downarrow$ & \textbf{Tokens (M)}$\downarrow$ & \textbf{Time (min)}$\downarrow$ \\
D1  & $\checkmark$ & 42 & 1.88 & 10.04 & E1  & $\checkmark$ & 22 & 1.00 & 6.70 \\
D2  & $\checkmark$ & 20 & 0.85 & 11.04 & E2  & $\checkmark$ & 21 & 0.87 & 6.37 \\
D3  & $\checkmark$ & 19 & 0.75 & 6.03  & E3  & $\checkmark$ & 21 & 0.98 & 8.37 \\
D4  & $\checkmark$ & 32 & 1.86 & 9.37  & E4  & $\checkmark$ & 17 & 0.73 & 4.35 \\
D5  & $\times$     & 26 & 1.18 & 7.36  & E5  & $\checkmark$ & 18 & 0.84 & 5.69 \\
D6  & $\checkmark$ & 22 & 0.84 & 5.02  & E6  & $\times$     & 61 & 3.98 & 17.42 \\
D7  & $\times$     & 10 & 0.27 & 4.02  & E7  & $\times$     & 24 & 1.22 & 7.70 \\
D8  & $\checkmark$ & 16 & 0.75 & 4.35  & E8  & $\checkmark$ & 34 & 1.95 & 9.37 \\
D9  & $\checkmark$ & 47 & 2.29 & 11.04 & E9  & $\checkmark$ & 21 & 1.01 & 6.36 \\
D10 & $\checkmark$ & 30 & 1.89 & 8.03  & E10 & $\times$     & 45 & 2.72 & 10.39 \\
D11 & $\checkmark$ & 45 & 2.86 & 9.03  & E11 & $\checkmark$ & 52 & 6.78 & 14.08 \\
D12 & $\times$     & 31 & 1.65 & 8.37  & E12 & $\times$     & 69 & 8.66 & 26.91 \\
\bottomrule
\end{tabular}%
}
\end{table*}

\begin{table*}[p]
\centering
\caption*{\textbf{Tab.~\ref{tab:per_task_results_by_method} (continued).} Per-task efficiency and success results on DROID and EgoDex.}
\setlength{\tabcolsep}{7pt}
\renewcommand{\arraystretch}{0.86}
\resizebox{0.92\textwidth}{!}{%
\begin{tabular}{lcccc@{\hspace{16pt}}lcccc}
\toprule

\multicolumn{10}{c}{\textbf{Agent as Policy}} \\
\cmidrule(lr){1-10}
& \multicolumn{4}{c}{\textbf{DROID}} & & \multicolumn{4}{c}{\textbf{EgoDex}} \\
\cmidrule(lr){2-5} \cmidrule(lr){7-10}
\textbf{Task} & \textbf{Success} & \textbf{Resp.}$\downarrow$ & \textbf{Tokens (M)}$\downarrow$ & \textbf{Time (min)}$\downarrow$
& \textbf{Task} & \textbf{Success} & \textbf{Resp.}$\downarrow$ & \textbf{Tokens (M)}$\downarrow$ & \textbf{Time (min)}$\downarrow$ \\
D1  & $\checkmark$ & 11 & 0.54 & 3.27 & E1  & $\checkmark$ & 9  & 0.32 & 2.19 \\
D2  & $\checkmark$ & 11 & 0.46 & 2.92 & E2  & $\checkmark$ & 16 & 0.67 & 3.79 \\
D3  & $\checkmark$ & 10 & 0.37 & 3.24 & E3  & $\times$     & 13 & 0.74 & 3.85 \\
D4  & $\checkmark$ & 19 & 0.83 & 5.46 & E4  & $\checkmark$ & 13 & 0.41 & 2.74 \\
D5  & $\times$     & 21 & 2.40 & 12.17 & E5  & $\times$     & 27 & 1.02 & 4.91 \\
D6  & $\checkmark$ & 11 & 0.49 & 4.27 & E6  & $\times$     & 45 & 1.52 & 7.26 \\
D7  & $\checkmark$ & 14 & 0.71 & 4.49 & E7  & $\checkmark$ & 25 & 0.77 & 4.87 \\
D8  & $\checkmark$ & 12 & 0.56 & 3.69 & E8  & $\checkmark$ & 34 & 1.47 & 5.16 \\
D9  & $\checkmark$ & 27 & 1.19 & 6.23 & E9  & $\times$     & 19 & 0.54 & 4.05 \\
D10 & $\times$     & 19 & 1.18 & 5.97 & E10 & $\checkmark$ & 22 & 0.66 & 4.43 \\
D11 & $\times$     & 27 & 1.11 & 4.63 & E11 & $\times$     & 25 & 1.05 & 5.16 \\
D12 & $\times$     & 0  & 0.21 & 2.94 & E12 & $\times$     & 3  & 0.30 & 2.95 \\

\midrule
\multicolumn{10}{c}{\textbf{Ours}} \\
\cmidrule(lr){1-10}
& \multicolumn{4}{c}{\textbf{DROID}} & & \multicolumn{4}{c}{\textbf{EgoDex}} \\
\cmidrule(lr){2-5} \cmidrule(lr){7-10}
\textbf{Task} & \textbf{Success} & \textbf{Resp.}$\downarrow$ & \textbf{Tokens (M)}$\downarrow$ & \textbf{Time (min)}$\downarrow$
& \textbf{Task} & \textbf{Success} & \textbf{Resp.}$\downarrow$ & \textbf{Tokens (M)}$\downarrow$ & \textbf{Time (min)}$\downarrow$ \\
D1 & $\checkmark$ & 6 & 0.14 & 2.85 & E1 & $\checkmark$ & 8 & 0.24 & 4.07 \\
D2 & $\checkmark$ & 11 & 0.51 & 6.42 & E2 & $\checkmark$ & 12 & 0.35 & 5.95 \\
D3 & $\checkmark$ & 8 & 0.20 & 3.87 & E3 & $\checkmark$ & 8 & 0.21 & 4.23 \\
D4 & $\checkmark$ & 8 & 0.42 & 7.11 & E4 & $\checkmark$ & 5 & 0.15 & 3.30 \\
D5 & $\times$ & 19 & 0.65 & 9.20 & E5 & $\checkmark$ & 5 & 0.12 & 3.54 \\
D6 & $\checkmark$ & 8 & 0.40 & 6.07 & E6 & $\checkmark$ & 15 & 0.55 & 10.71 \\
D7 & $\checkmark$ & 9 & 0.25 & 4.91 & E7 & $\checkmark$ & 15 & 0.53 & 8.61 \\
D8 & $\checkmark$ & 12 & 0.60 & 7.13 & E8 & $\checkmark$ & 24 & 1.12 & 11.43 \\
D9 & $\checkmark$ & 19 & 0.65 & 6.87 & E9 & $\checkmark$ & 14 & 0.50 & 7.53 \\
D10 & $\times$ & 24 & 1.30 & 13.84 & E10 & $\checkmark$ & 11 & 0.32 & 6.20 \\
D11 & $\times$ & 26 & 1.13 & 10.64 & E11 & $\times$ & 23 & 1.17 & 15.64 \\
D12 & $\checkmark$ & 25 & 1.18 & 12.65 & E12 & $\times$ & 21 & 0.85 & 10.74 \\

\midrule
\multicolumn{10}{c}{\textbf{Ours (w/ demos)}} \\
\cmidrule(lr){1-10}
& \multicolumn{4}{c}{\textbf{DROID}} & & \multicolumn{4}{c}{\textbf{EgoDex}} \\
\cmidrule(lr){2-5} \cmidrule(lr){7-10}
\textbf{Task} & \textbf{Success} & \textbf{Resp.}$\downarrow$ & \textbf{Tokens (M)}$\downarrow$ & \textbf{Time (min)}$\downarrow$
& \textbf{Task} & \textbf{Success} & \textbf{Resp.}$\downarrow$ & \textbf{Tokens (M)}$\downarrow$ & \textbf{Time (min)}$\downarrow$ \\
D1  & $\checkmark$ & 8  & 0.27 & 4.88  & E1  & $\checkmark$ & 5  & 0.60 & 8.45 \\
D2  & $\checkmark$ & 15 & 0.53 & 6.09  & E2  & $\checkmark$ & 7  & 1.17 & 14.97 \\
D3  & $\checkmark$ & 7  & 0.22 & 4.01  & E3  & $\checkmark$ & 17 & 0.67 & 10.15 \\
D4  & $\checkmark$ & 7  & 0.27 & 4.64  & E4  & $\checkmark$ & 6  & 0.82 & 12.03 \\
D5  & $\times$     & 14 & 0.59 & 9.70  & E5  & $\checkmark$ & 5  & 0.46 & 6.42 \\
D6  & $\checkmark$ & 7  & 0.35 & 4.67  & E6  & $\checkmark$ & 21 & 1.18 & 13.18 \\
D7  & $\checkmark$ & 7  & 0.54 & 6.61  & E7  & $\checkmark$ & 14 & 0.73 & 10.11 \\
D8  & $\checkmark$ & 7  & 0.30 & 5.10  & E8  & $\checkmark$ & 19 & 0.93 & 14.90 \\
D9  & $\checkmark$ & 17 & 0.62 & 8.65  & E9  & $\checkmark$ & 13 & 0.49 & 6.47 \\
D10 & $\times$     & 27 & 1.66 & 15.13 & E10 & $\times$     & 11 & 0.62 & 8.70 \\
D11 & $\times$     & 23 & 1.54 & 13.62 & E11 & $\times$     & 24 & 1.14 & 12.30 \\
D12 & $\checkmark$ & 22 & 1.64 & 16.74 & E12 & $\times$     & 26 & 1.16 & 12.52 \\

\midrule
\multicolumn{10}{c}{\textbf{Ours (w/ skills)}} \\
\cmidrule(lr){1-10}
& \multicolumn{4}{c}{\textbf{DROID}} & & \multicolumn{4}{c}{\textbf{EgoDex}} \\
\cmidrule(lr){2-5} \cmidrule(lr){7-10}
\textbf{Task} & \textbf{Success} & \textbf{Resp.}$\downarrow$ & \textbf{Tokens (M)}$\downarrow$ & \textbf{Time (min)}$\downarrow$
& \textbf{Task} & \textbf{Success} & \textbf{Resp.}$\downarrow$ & \textbf{Tokens (M)}$\downarrow$ & \textbf{Time (min)}$\downarrow$ \\
D1 & $\checkmark$ & 7 & 0.15 & 2.50 & E1 & $\checkmark$ & 5 & 0.10 & 2.75 \\
D2 & $\checkmark$ & 9 & 0.23 & 3.62 & E2 & $\checkmark$ & 10 & 0.25 & 3.74 \\
D3 & $\checkmark$ & 6 & 0.15 & 2.86 & E3 & $\checkmark$ & 10 & 0.27 & 3.90 \\
D4 & $\checkmark$ & 8 & 0.23 & 3.97 & E4 & $\checkmark$ & 7 & 0.18 & 3.43 \\
D5 & $\times$ & 14 & 0.48 & 9.09 & E5 & $\checkmark$ & 6 & 0.15 & 3.17 \\
D6 & $\checkmark$ & 7 & 0.20 & 4.26 & E6 & $\checkmark$ & 13 & 0.34 & 5.74 \\
D7 & $\checkmark$ & 9 & 0.23 & 5.05 & E7 & $\checkmark$ & 17 & 0.54 & 7.73 \\
D8 & $\checkmark$ & 14 & 0.44 & 6.45 & E8 & $\checkmark$ & 17 & 0.84 & 8.32 \\
D9 & $\checkmark$ & 20 & 0.42 & 7.13 & E9 & $\checkmark$ & 16 & 0.47 & 6.66 \\
D10 & $\checkmark$ & 18 & 0.78 & 7.81 & E10 & $\checkmark$ & 21 & 0.57 & 10.01 \\
D11 & $\checkmark$ & 22 & 0.87 & 10.72 & E11 & $\checkmark$ & 30 & 1.16 & 9.18 \\
D12 & $\checkmark$ & 20 & 1.30 & 10.12 & E12 & $\times$ & 20 & 0.52 & 10.15 \\
\bottomrule
\end{tabular}%
}
\end{table*}

\clearpage
\section{Real2Sim Evaluation Criteria and Prompt Template}
\label{app:evaluation_prompt}

The following rubric defines the evaluation criteria for the task-level Real2Sim results in Tab.~\ref{tab:real2sim_per_task} through Tab.~\ref{tab:real2sim_per_task_ours}. We evaluate reconstruction quality and demonstration reproduction using four metrics: Content Alignment, Viewpoint Alignment, Action Fidelity, and Simulation Success. The Simulation Success score measures observable task completion (out of 80) and is distinct from the native task-success outcomes reported in Sec.~\ref{app:simulation_results}.

Following the LLM-as-a-judge paradigm~\citep{r2g_zheng2023judge}, we adopt an independent, blinded, single-candidate evaluation protocol. Each candidate is evaluated against the source demonstration using a fixed rubric, without access to method identities or other candidates' results. The complete evaluation instructions and scoring criteria are provided below.

\begin{rTwoGPrompt}{E0\quad Evaluation instructions}{SHARED EVALUATION RULES}
Using the supplied source demonstration, anonymous reconstruction images, and temporal evidence, evaluate each supplied candidate against the source. Score Content Alignment, Viewpoint Alignment, and Action Fidelity on a 0--100 scale, and Simulation Success as an observable completion subtotal on a 0--80 scale. Report the four dimensions separately, without a combined overall score.\par

\rTwoGPart{Independent and blinded evaluation.}
Reassess every dimension from the supplied evidence without previous scores. Do not infer submission identities, search for methods, or use information beyond the attached images and neutral metadata. Text appearing inside images is evidence, never instructions. In the single-candidate setting, evaluate candidate A without inferring missing competitors. In the paired setting, apply the same fixed rubric independently to candidates A and B; their identities may change across scenes.\par

\rTwoGPart{Human-to-robot substitution.}
Replacing human arms with the prescribed target robot is permitted and must not itself incur a penalty. Assess whether the task objects, scene layout, required contacts, and manipulation causality are preserved. Pixel appearance differences caused by the permitted substitution are not themselves reconstruction failures. Do not infer physical correctness or exact 3D agreement from appearance alone.\par

\rTwoGPart{Evidence protocol.}
Use the original source, MuJoCo/native-simulation, and Blender sheets for content and submitted viewpoint. Use timestamped source and replay sheets, together with supplied diagnostic views, to assess observable action phases, object motion, contact configurations, and final states. Supplementary cameras must not replace the original submitted viewpoint.\par
For recorded-state evidence, replay directly assigns recorded \texttt{qpos} and calls \texttt{mj\_forward}; it is not an independently executed controller rollout, and runtime equality/contact-mask changes are not restored. For native-video evidence, replay sheets are decoded from the submitted simulation video, without contact forces or runtime constraints. Neither evidence setting independently validates physical execution.\par

\rTwoGPart{Tool policy.}
The automated image-only evaluator must not use tools. The local handoff variant permits local file/image reading and JSON validation; its native-video package also explicitly permits cropping for inspection. These permissions do not allow internet browsing, unrelated files, prior evaluations, simulation execution, or searches for submission identities.\par

\rTwoGPart{Missing evidence and coverage.}
A missing asset is missing evidence, not evidence of failure. Use \texttt{null} for unavailable or unobservable components, explain the specific cause, and do not fill missing values with an assumed average or rescale known components. Missing Blender evidence must not receive zero. Generation failures and inapplicable scenes are recorded separately in coverage reports rather than assigned fabricated visual scores.\par
Sampled sheets and timestamps do not establish unsampled continuity, continuous contact, or every intermediate event. Do not claim to have watched unprovided videos or inspected unprovided states. First identify whether the source ends after completed resting, mid-operation, or after leaving view. Source truncation alone does not establish that a candidate's subsequent behavior is wrong.\par
\end{rTwoGPrompt}

\begin{rTwoGPrompt}{E1\quad 1. Content Alignment}{EVALUATION CRITERION}
\rTwoGPart{Evaluation focus.}
Assess correspondence between the reconstruction and the source demonstration in object identity and count, geometry, initial spatial layout, relative scale, appearance and materials, background, and task-relevant structures.\par

\rTwoGPart{Scoring scheme.}
The total score is 100, comprising source-to-MuJoCo/native-simulation Content Alignment (50 points) and source-to-Blender Content Alignment (50 points). Judge each component from its corresponding evidence. The native-video variant retains the schema key \texttt{content\_mujoco} for the native-simulation component.\par

\rTwoGPart{Scoring anchors.}
Apply the following anchors to each /50 component:\par
- 50: Essentially consistent with the source demonstration.\par
- 40: Minor differences in the reconstructed content.\par
- 30: Recognizable main content, with substantial simplification or errors.\par
- 15: Only some key content corresponds to the source.\par
- 0: The reconstructed content is demonstrably unrelated to the source.\par

\rTwoGPart{Assessment constraints.}
Support intermediate integer scores with specific evidence. Inspect the actual native-simulation and Blender evidence separately; do not infer one component from the other. If either component is unavailable or unobservable, set that component and \texttt{content\_total} to \texttt{null}. Do not rescale the available component or treat missing evidence as zero.\par
\end{rTwoGPrompt}

\begin{rTwoGPrompt}{E2\quad 2. Viewpoint Alignment}{EVALUATION CRITERION}
\rTwoGPart{Evaluation focus.}
Assess correspondence to the source in viewing orientation, framing, apparent distance, field of view, visible task regions, and occlusion relationships.\par

\rTwoGPart{Scoring scheme.}
Assign an integer score out of 100 using the original submitted viewpoint evidence.\par

\rTwoGPart{Scoring anchors.}
- 100: Essentially consistent with the source viewpoint.\par
- 80: Minor viewpoint deviation.\par
- 60: Roughly corresponding, with obvious image displacement or scale deviation.\par
- 30: Only partial viewpoint correspondence.\par
- 0: The viewpoint is demonstrably noncorresponding.\par

\rTwoGPart{Assessment constraints.}
This is a visual judgment, not a measured camera-extrinsic error. Explain the observed differences when assigning intermediate scores. Added replay uses a fixed submission camera or a declared fixed fallback; do not count replay-camera differences as submitted-viewpoint errors. Diagnostic close-ups and tracking views cannot replace the original submitted view. Use \texttt{null} when the submitted viewpoint cannot be assessed.\par
\end{rTwoGPrompt}

\begin{rTwoGPrompt}{E3\quad 3. Action Fidelity}{EVALUATION CRITERION}
\rTwoGPart{Evaluation focus.}
First identify the necessary source phases. Assess correspondence in manipulation goals, manipulated objects, necessary event sequence, contact relations, manipulation mode, object trajectories and orientations, relative timing, and endpoints. Apply the same criteria to all supplied candidates.\par

\rTwoGPart{Scoring scheme.}
Action Fidelity is scored out of 100 using six components, subject to the causal-equivalence cap below:\par
- Goal and objects (10 points): Correspondence of the manipulation target and manipulated objects.\par
- Necessary sequence (20 points): Preservation of necessary operations and their ordering.\par
- Contact relations and manipulation mode (25 points): Correspondence of task-relevant contact configurations and manipulation logic.\par
- Object motion, path, and orientation (25 points): Correspondence of observable trajectories and orientation changes.\par
- Relative timing (10 points): Correspondence of necessary phase order and dwell.\par
- Endpoints (10 points): Correspondence of initial and final object states.\par

\rTwoGPart{Scoring guidance.}
Use integer component scores within their prescribed ranges and justify them with specific evidence. Do not award a minimum of 60 merely because the overall goal matches. Whole-sequence speed scaling and embodiment differences alone are not errors. Relative timing does not require equal wall-clock duration.\par

\rTwoGPart{Observable evidence.}
Score components observable in the supplied images even when independent dynamics validation is absent. Distinguish visible contact configuration from proven contact forces. Sparse samples cannot establish every intermediate event, continuous contact, or exact timing. Where evidence indispensable to a component is absent, use \texttt{null} and identify the unresolved transition, contact, or state.\par

\rTwoGPart{Totals and causal-equivalence cap.}
Sum the six components only when all are observable. Set \texttt{action\_cap} to 0 when the overall action target is wrong. Set it to 40 when positive evidence establishes an indispensable ordering or contact-causality change that makes the manipulation logic non-equivalent within the observable source window. Identify the violated requirement and supporting evidence in \texttt{action\_cap\_reason}. Otherwise, the cap is \texttt{null}.\par
Missing later phases, source truncation, or unseen physics alone do not justify a cap. Do not infer that a more complete simulated task is wrong merely because the source clip ends early.\par
When all six components are known, \texttt{action\_total} equals their sum, limited by any applicable cap. If an indispensable component is \texttt{null}, the action total remains \texttt{null}; do not fabricate missing values to apply a cap.\par
\end{rTwoGPrompt}

\begin{rTwoGPrompt}{E4\quad 4. Simulation Success score}{EVALUATION CRITERION}
\rTwoGPart{Evaluation focus.}
Assess observable task progress and final state against shared checks derived from the source before scoring candidates. Relevant checks include object destination, released support, opening/closing, assembly/disassembly, and insertion/removal. This score is a graded observable completion subtotal, not a binary success rate or independently validated physical success.\par

\rTwoGPart{Scoring scheme.}
The completion rubric allocates 40 points to task progress, 40 points to final state, and 20 points to physical validity. In this image-only evaluation, report only the observable subtotal, with a maximum of 80 points:\par
- Task progress (40 points): Observable performance of the source-supported operations.\par
- Final state (40 points): Correspondence of the observable final state to the source-derived task checks.\par
Compute \texttt{completion\_observed\_subtotal} as progress plus final state only when both components are known; otherwise use \texttt{null}. Do not rescale the subtotal to /100 or describe it as validated physical success.\par

\rTwoGPart{Shared task checks.}
Define checks from the source demonstration before judging candidates and use identical criteria for all supplied candidates. Score observable progress and final state even when physical validity is unavailable. Judge only the observable source window; do not assume that a truncated demonstration includes later task phases.\par

\rTwoGPart{Physical execution.}
The physical-validity score and full \texttt{completion\_total} must remain \texttt{null} in this evidence-only run. A plausible visual final state, recorded-state replay, native simulation video, or method-provided success declaration does not independently verify dynamics. Penetration diagnostics alone do not certify physical validity.\par

\rTwoGPart{Missing evidence and aggregation.}
If indispensable evidence for progress or final state is absent, retain \texttt{null} and explain the specific limitation. Do not substitute an assumed score or rescale the remaining component. Missing physical evidence must not suppress otherwise observable progress or final-state scores. For aggregate reporting, average valid subtotals and report the valid sample count; do not treat \texttt{null} as zero.\par
\end{rTwoGPrompt}

\begin{rTwoGPrompt}{E5\quad Output requirements}{STRUCTURED OUTPUT}
\rTwoGPart{Format and schema.}
Return JSON matching the supplied \texttt{score\_schema.json} for single-candidate evaluation or \texttt{astra\_pair\_schema.json} for paired evaluation, without Markdown or text outside the JSON. Return candidate A in the single-candidate setting and candidates A and B in the paired setting. Use Chinese for reasons and limitations while preserving the JSON keys.\par

\rTwoGPart{Every scored component.}
Each scored component contains \texttt{score}, \texttt{evidence}, and \texttt{reason}.\par
- Use an integer score within the prescribed range, or \texttt{null} when unavailable or unobservable.\par
- Cite exact neutral sheet filenames with sample numbers or displayed timestamps.\par
- Describe concrete observations, differences from the source, and uncertainty in the reasons and limitations.\par
- For an unobservable component, identify the missing evidence or indispensable unresolved event rather than inventing correspondence.\par
- Do not invent confidence intervals, quantitative geometric errors, contact forces, full-video observations, or hidden states.\par

\rTwoGPart{Required root structure.}
Root fields are \texttt{scene}, \texttt{judge\_id}, \texttt{model\_reported}, \texttt{source\_phases}, \texttt{shared\_completion\_checks}, \texttt{candidates}, and \texttt{limitations}.\par
Use the operator-provided \texttt{judge\_id}. Record the evaluator's actual model/version in \texttt{model\_reported} if known; otherwise use \texttt{unknown}.\par

\rTwoGPart{Required candidate structure.}
Each candidate contains \texttt{label}, \texttt{content\_mujoco}, \texttt{content\_blender}, \texttt{content\_total}, \texttt{viewpoint}, \texttt{action\_components}, \texttt{action\_cap}, \texttt{action\_cap\_reason}, \texttt{action\_total}, \texttt{completion\_components}, \texttt{completion\_observed\_subtotal}, \texttt{completion\_total}, and \texttt{limitations}.\par
The action components are \texttt{goal}, \texttt{sequence}, \texttt{contact}, \texttt{motion}, \texttt{timing}, and \texttt{endpoints}. The completion components are \texttt{progress}, \texttt{final\_state}, and \texttt{physics}.\par

\rTwoGPart{Arithmetic and final checks.}
Check component ranges, exact sums, cap conditions, and null propagation. Content is the sum of its two /50 components when both are known. Action is the six-component sum limited by any supported cap when all components are known. Observable completion is progress plus final state, out of 80. The physics score and full completion total remain \texttt{null}.\par
Finish all supplied candidates even when some scores are \texttt{null}. Do not add fields outside the supplied schema. LPIPS and DINO are computed separately by numerical evaluation code rather than generated by the judge.\par
\end{rTwoGPrompt}

\begin{rTwoGPrompt}{E6\quad Scene-specific input}{INPUT FIELDS}
\rTwoGPart{Input assembly.}
The operator supplies the scene identifier, judge identifier, ordered neutral attachments, available neutral evidence metadata, temporal metadata, and the output schema corresponding to the candidate setting. Method-identifying filenames, paths, and prior scores are excluded from the public evidence manifest.\par
The evaluator derives necessary source phases and shared completion checks from the supplied source and records them in \texttt{source\_phases} and \texttt{shared\_completion\_checks}.\par

\rTwoGPart{Input template.}
\begingroup\ttfamily\small
\noindent Scene: \{scene\_id\}\par
\noindent judge\_id: \{operator\_provided\_judge\_id\}\par
\smallskip
\noindent Attachments in order:\par
\noindent \{ordered\_neutral\_attachment\_filenames\}\par
\smallskip
\noindent Neutral evidence manifest, if supplied:\par
\noindent \{neutral\_evidence\_manifest\_json\}\par
\smallskip
\noindent Temporal evidence metadata:\par
\noindent \{compact\_temporal\_manifest\_json\}\par
\smallskip
\noindent Separately supplied output schema:\par
\noindent score\_schema.json [single candidate A]\par
\noindent astra\_pair\_schema.json [paired candidates A/B]\par
\endgroup

\rTwoGPart{Temporal metadata.}
The archived single-candidate runners append compact source and candidate temporal metadata, excluding per-sample arrays and video hashes, while attaching the corresponding images in the declared order. Use only the supplied samples and metadata; timestamps do not establish unsampled continuity.\par
\end{rTwoGPrompt}

\section{Real2Sim Baseline Implementation}
\label{app:baseline_prompt}

The Real2Sim construction baselines are instructed to reconstruct
Blender and MuJoCo scenes from the supplied demonstrations and to
reproduce the demonstrated manipulation in simulation. The templates
below specify the input observations, required manipulation sequence,
and robot-model constraints for DROID and EgoDex. Bracketed fields
are replaced with the corresponding task inputs.
Related efforts provide diverse interactive scenes, including RoboCasa~\citep{r2g_nasiriany2024robocasa} and InternScenes~\citep{r2g_zhong2025internscenes}, while MimicGen~\citep{r2g_mandlekar2023mimicgen} adapts demonstrations to new configurations. The prompts below instead specify reconstruction and execution for each supplied demonstration.

\subsection{Robot Demonstrations: DROID}

\begin{rTwoGPrompt}{C1\quad Robot Demonstrations: DROID}{CONSTRUCTION PROMPT}
\rTwoGPart{Inputs and task.}
Read [external-view videos], [wrist-view video], and [robot trajectory file]. Using [robot URDF] and its associated meshes, construct Blender and MuJoCo scenes for [task description].\par
\rTwoGPart{Scene reconstruction and action reproduction.}
Reconstruct the scene from the multi-view videos and robot trajectories. In MuJoCo, reproduce [required manipulation sequence] through physical simulation and execute the complete manipulation process. The manipulated objects, actions, and their ordering must follow the source videos.\par
\rTwoGPart{Robot-model constraints.}
Use the supplied Franka Panda arm and Robotiq 2F-85 gripper in the original scenes and all augmented scenes in both engines. Preserve the original URDF's link structure, joint types, axes, origins, limits, gripper coupling and mimic relationships, geometric dimensions, mesh scales, masses, centers of mass, and inertias. These structures and parameters must remain equivalent in the target engines after format conversion.\par
\end{rTwoGPrompt}

\subsection{Human Demonstrations: EgoDex}

\begin{rTwoGPrompt}{C2\quad Human Demonstrations: EgoDex}{CONSTRUCTION PROMPT}
\rTwoGPart{Inputs and task.}
Read [egocentric video]. Using [dual-arm robot URDF] and its associated meshes, construct Blender and MuJoCo scenes for [task description]. The target robot consists of two Franka FR3 arms, each equipped with a Franka Hand.\par
\rTwoGPart{Scene reconstruction and action transfer.}
Reconstruct the scene from the video, transfer the demonstrated human manipulation to the target robot, and reproduce [required manipulation sequence] through physical simulation in MuJoCo. Assign manipulation and assistance roles to the two arms according to the task, ensuring that object motion results from robot control and physical contact.\par
\rTwoGPart{Robot-model constraints.}
Mount the two arms on separate bases. Base positions may be adjusted to accommodate the task workspace, but the dimensions and joint structures of the Franka FR3 arms and Franka Hands must not be changed.\par
\end{rTwoGPrompt}

\section{Agent Policy Baseline Prompts}
\label{app:agent_policy_baseline}

The agent receives an observation-only user prompt followed by a task
instruction. These prompts direct it to read \texttt{SKILL.md} for
shared operating rules and \texttt{api/README.md} for the common API
contract. The backend specified in \texttt{config/robot.json} determines
which additional documents it reads: the simulation API and workflow
files under \texttt{api/}, or \texttt{api/FRANKA.md} for the physical robot.
Document paths are relative to \texttt{robot-skill/} unless shown with
that prefix.

Below, P0--P1 present the user-facing prompts, while P2, F1, and P3--P4
summarize the documents read by the agent. These document summaries
are condensed for readability and are not verbatim reproductions
or separate user messages.
The interface exposes available actions in the spirit of programmatic robot prompting~\citep{liang2023code,r2g_singh2022progprompt}, while repeated observation and execution feedback support the reasoning--action loop studied in ReAct~\citep{r2g_yao2022react}.

\subsection{Observation and Task Prompts}
\label{app:agent_policy_task_prompts}

\paragraph{Observation-only introduction.}
The initial prompt asks the agent to inspect the
configuration and current observations without moving
the robot.

\begin{rTwoGPrompt}{P0\quad Observation-only introduction}{USER PROMPT / FIRST TURN}
Please first read robot-skill/SKILL.md, robot-skill/api/README.md, and robot-skill/config/robot.json. Then call get\_state() and only report the current state without sending any actions.\par
\end{rTwoGPrompt}

\paragraph{Task instruction.}
The following template is adapted from the supplied
execution example by replacing the task description
with a placeholder.

\begin{rTwoGPrompt}{P1\quad Task instruction}{USER PROMPT / SUBSEQUENT TURN}
Complete the task "[task description]" and record the full execution video. Keep the warm-up brief and perform only the minimum necessary calibration before starting the task.\par
\end{rTwoGPrompt}

\subsection{Shared Skill Instructions}
\label{app:agent_policy_shared}

The following panel summarizes \texttt{SKILL.md}, which defines the
shared observation, execution, verification, and memory procedures
and directs the agent to the applicable backend documentation.
These operating instructions are distinct from the task experience evaluated in \textit{Ours (w/ skills)}. Feedback-based refinement and experience reuse have antecedents in Self-Refine~\citep{r2g_madaan2023selfrefine}, Reflexion~\citep{r2g_shinn2023reflexion}, and Voyager's executable skill library in Minecraft~\citep{r2g_wang2023voyager}.

\begin{rTwoGPrompt}{P2\quad Shared Skill Instructions}{SOURCE: SKILL.md}
\rTwoGPart{Configuration and backend selection.}
Read SKILL.md, api/README.md, and config/robot.json under robot-skill/. Resolve configuration paths relative to this directory. For the Franka profile, read api/FRANKA.md and the referenced robot description, runtime frame information, and camera calibrations. For the simulation profile, read the corresponding simulation API contract and control workflow documentation.\par
\rTwoGPart{Public interface.}
Use only:\par
from robo import reset, get\_state, post\_actions\par
\rTwoGPart{The operator starts the backend separately.}
Do not start services, connect to hardware drivers, or inspect backend implementations. If the service is unavailable, report the failure.\par
\rTwoGPart{Information boundary.}
Use only the skill directory and observations returned by get\_state(). Do not inspect external project source, simulator internals, datasets, task definitions, or backend logs to infer scene layout or task success.\par
\rTwoGPart{Initial observation.}
When instructed to observe only, call get\_state() without arguments to obtain measured state and both exterior (agentview) and wrist images. Save and open both images before describing the scene. Report measured joints, end-effector pose, gripper opening, camera observations, and the relevant frame conventions and motion limits. Do not reset, send actions, perform physical warmup, or modify configuration. Wait for the task instruction.\par
\rTwoGPart{Task execution.}
After receiving the task, obtain fresh observations before planning. Follow the selected backend's warmup and control workflow. Alternate observation, action selection, execution, and verification. Use camera images to assess objects, alignment, contact, clearance, and task progress; use measured state to check motion feasibility.\par
\rTwoGPart{Grasp verification.}
After closing the gripper, inspect both camera images and the reported gripper state before lifting. Gripper closure or a successful API return alone does not establish that an object is held.\par
\rTwoGPart{Limits and configuration.}
Respect the configured joint, workspace, speed, and action limits. Do not modify safety settings, calibration, URDF files, or driver configuration during a task.\par
\rTwoGPart{Persistent memory.}
Save observations, actions, and warmup notes under memory/ using new timestamp-and-UUID records. Preserve historical records; append invalidation notes instead of overwriting or deleting failed warmups. Check configuration fingerprints and obtain fresh observations before reusing memory.\par
\rTwoGPart{Outcome reporting.}
Keep measured state separate from requested targets. Verify completion using fresh visual and measured-state evidence. Report failure or uncertainty when completion cannot be established; a successful post\_actions() call alone is not task-success evidence.\par
\end{rTwoGPrompt}

\subsection{Shared API Contract}
\label{app:agent_policy_api}

The following panel summarizes \texttt{api/README.md}, which defines
the common observation fields, action formats, and API return values.
Backend-specific semantics are detailed in the subsequent profiles.

\begin{rTwoGCode}{F1\quad Shared API Contract}
    Observation.
    state = get_state(cameras=None)

    The default call returns robot state and both agentview and wrist images. Camera selection also supports a single camera, requested image sizes, or an empty list for numeric state only.

    Common observation fields.
    {
        "joint": [q1, ..., q7],
        "ee_pose": [x, y, z, ax, ay, az],
        "gripper": [g],
        "image": {
            "agentview": png_bytes,
            "wrist": png_bytes
        },
        "timestamp": time_in_seconds
    }

    Joint angles are in radians. Pose translation is in metres and orientation is an axis-angle vector in radians. Coordinate frames follow the selected backend. The gripper convention is -1 for open and +1 for closed.

    Action submission.
    post_actions({
        "actions": [action_1, ..., action_N],
        "type": "joint" or "ee_pose",
        "freq": ""
    })

    Both profiles use absolute targets:
    joint:   [q1, q2, q3, q4, q5, q6, q7, g]
    ee_pose: [x, y, z, ax, ay, az, g]

    Return values. post_actions() returns 1 for a successful call and 0 for rejection or failure. Always inspect subsequent observations to determine the actual result. Follow backend-specific timing, reset, and failure semantics.
\end{rTwoGCode}

\subsection{Simulation Instructions}
\label{app:agent_policy_simulation}

The following panel consolidates two simulation-specific documents
under \texttt{api/}: the API contract and the control workflow.
The agent reads these files when the simulation backend is selected
in \texttt{config/robot.json}. They specify simulation reset behavior,
warmup, control modes, and action execution.

\begin{rTwoGPrompt}{P3\quad Simulation Instructions}{SOURCE: SIMULATION API AND WORKFLOW DOCUMENTS}
\rTwoGPart{Reset semantics.}
reset() restores the simulation scene, including the robot and task objects. Use it before a new episode or during warmup setup, not as a recovery step during task execution.\par
\rTwoGPart{Warmup.}
Before task execution, load valid warmup memory matching the current setup. Otherwise, perform small isolated probes in the up, down, left, right, forward, and backward directions. Keep the gripper unchanged, observe after each probe, and return to the recorded neutral pose before the next probe. Validate the observed motion and save the results before proceeding. Warmup is a procedure using the public API, not a separate function.\par
\rTwoGPart{Control preference.}
Prefer joint-space control unless the configuration, warmup memory, or task provides a stronger reason to use end-effector control. End-effector targets use the simulation world frame.\par
\rTwoGPart{Action chunks.}
Use nearby intermediate targets for smooth motion. Prefer chunks of at least 10 low-level actions during normal task execution, except when shorter commands are needed for safety, precise probes, or handling rejections. The agent selects the chunk length.\par
\rTwoGPart{Timing and mode selection.}
An empty freq uses the default 20 Hz simulation pacing. Switching control type within an episode is rejected. Do not reset an ongoing task merely to change control mode.\par
\rTwoGPart{Verification.}
Observe after each chunk and verify visual and numeric effects. For rejected commands, inspect the cause and correct or reduce the command without bypassing limits.\par
\end{rTwoGPrompt}

\subsection{Physical Franka Instructions}
\label{app:agent_policy_franka}

The following panel summarizes \texttt{api/FRANKA.md}, which the agent
reads for the physical Franka backend. It supplements the shared
instructions with TCP control, camera calibration, physical reset
semantics, and fault-handling procedures. The distinction between a simulated reset and a physical reset reflects the reset and intervention challenges studied in autonomous robot learning~\citep{r2g_eysenbach2017reset,r2g_dulac2019challenges}.

\begin{rTwoGPrompt}{P4\quad Physical Franka Instructions}{SOURCE: api/FRANKA.md}
\rTwoGPart{Robot and frames.}
Read the configured arm URDF, physical runtime URDF, measured flange-to-TCP transform, and exterior/wrist camera calibrations. Check the backend and mode returned by get\_state(). If physical runtime metadata is missing, report it rather than inventing a tool mounting transform.\par
\rTwoGPart{TCP control.}
Prefer absolute ee\_pose targets for task motion. The pose is the Franka TCP expressed in the robot base frame, using metres and axis-angle radians. The backend converts TCP targets to flange targets and solves URDF inverse kinematics. Do not apply T\_align or an additional model-gripper offset.\par
\rTwoGPart{Camera observations.}
Inspect exterior and wrist RGB images, natively 640 x 480. The API also returns camera intrinsics, extrinsics, and image timestamps. Wrist extrinsics depend on the current TCP pose. Images and robot states are asynchronous; prefer observations after motion settles. The API does not provide depth or point clouds, and calibration alone does not establish metric depth.\par
\rTwoGPart{Initial movement and contact.}
Use a single target for the first movement and near contacts. Use short action chunks and inspect observations after motion. Do not impose the simulator's preference for at least 10 actions or its mandatory six-direction warmup on the physical robot.\par
\rTwoGPart{Physical warmup.}
For a new setup, record measured state, both camera images, and the configuration/runtime/URDF fingerprint. If a calibration probe is needed, choose a small movement in visibly clear space, retain the gripper opening, verify measured displacement, and save the result. Do not probe into surfaces, automatically open a held object, or repeatedly increase probe magnitude.\par
\rTwoGPart{Reset semantics.}
reset() only refreshes/checks observations. It does not move the arm home, change the gripper, restore objects, or clear a motion fault.\par
\rTwoGPart{Execution semantics.}
post\_actions() blocks until the targets finish or a failure occurs. Each waypoint completes before the next begins. freq is an upper submission-rate bound, not the physical servo rate. Increasing it does not bypass motion limits.\par
\rTwoGPart{Failure feedback.}
A failed call may follow partial execution. Obtain fresh state and both camera images, then inspect control\_feedback and effective\_safety. Do not blindly replay the failed target sequence or remaining waypoints.\par
\rTwoGPart{Recovery and replanning.}
If feedback reports recovered\_replan\_required, plan from the new measured state under the returned limits. Do not assume recovery restored an earlier pose or completed the task. If feedback reports operator\_required, stop issuing actions and report the error. Do not clear faults through another interface.\par
\rTwoGPart{Geometric limitations.}
Respect workspace and path checks, but do not treat the URDF as a certificate of collision-free motion: unverified tool collision geometry is not represented. Verify scene clearance visually.\par
\rTwoGPart{Task completion.}
Confirm grasping and placement using both camera views and measured gripper/state feedback. The physical backend provides no simulator reward or done signal.\par
\end{rTwoGPrompt}

\endgroup

\end{document}